\documentclass[10pt,twocolumn,letterpaper]{article}

\usepackage[pagenumbers]{cvpr} % To force page numbers, e.g. for an arXiv version

\usepackage{float}
\usepackage{pifont}
\usepackage{footnote}
\usepackage{enumitem}
\usepackage{bm}
\usepackage{arydshln}
\usepackage{booktabs}
\usepackage{multicol}
\usepackage{multirow}
\usepackage{color}
\usepackage{xcolor}     
\usepackage{colortbl}
\usepackage{soul}
\usepackage{bbding}
\usepackage{makecell}
\usepackage{mathtools}
\usepackage{imakeidx}
\usepackage{setspace}
\usepackage{threeparttable}
\makeindex
\usepackage{arydshln}
\usepackage{lipsum}
\usepackage{natbib}
\usepackage[toc]{multitoc}
\usepackage[edges]{forest}
\usepackage[normalem]{ulem}

\usepackage{bbding}
\usepackage[most]{tcolorbox}

\usepackage{algorithm}
\usepackage{algorithmic}

\usepackage{minitoc}
\usepackage[toc,page,header]{appendix}

\definecolor{orchid}{rgb}{0.85, 0.44, 0.84}
\definecolor{rubinred}{rgb}{0.82, 0.0, 0.28}
\definecolor{flagship}{rgb}{0.93, 0.06, 0.41}
\definecolor{radiologist}{rgb}{0.50, 0.50, 1}

\newcommand{\method}{{\fontfamily{ppl}\selectfont
ScaleMAI}}
\newcommand{\flagship}{{\fontfamily{ppl}\selectfont
Flagship Model}}
\newcommand{\dataset}{{\fontfamily{ppl}\selectfont
PanTS-XL}}

\newcommand{\yes}{\textcolor{flagship}{\ding{52}}} 
\newcommand{\no}{\textcolor{lightgray}{\ding{55}}}

\newcommand{\numofct}{47,315}
\newcommand{\numofnewct}{12,000}
\newcommand{\numofclass}{88}
\newcommand{\numofmask}{4,163,720}
\newcommand{\numofcountry}{13}
\newcommand{\numofhospital}{112}

\newcolumntype{P}[1]{>{\centering\arraybackslash}p{#1}}
\newlength\savewidth

\definecolor{cvprblue}{rgb}{0.21,0.49,0.74}
\usepackage[pagebackref,breaklinks,colorlinks,allcolors=cvprblue]{hyperref}
\usepackage[normalem]{ulem} % for Strikeout

\def\paperID{200} % *** Enter the Paper ID here
\def\confName{CVPR}
\def\confYear{2026}

\title{Expectation-Maximization as the Engine of Scalable Medical Intelligence}
\author{
Wenxuan Li\textsuperscript{1} \quad 
Pedro R. A. S. Bassi\textsuperscript{1,2,3} \quad
Tianyu Lin\textsuperscript{1} \quad
Yu-Cheng Chou\textsuperscript{1} \quad
Jakob Wasserthal\textsuperscript{4} \\
Xinze Zhou\textsuperscript{1} \quad
Qi Chen\textsuperscript{1} \quad
Fabian Isensee\textsuperscript{5} \quad
Yannick Kirchhoff\textsuperscript{5} \quad
Maximilian Rokuss\textsuperscript{5} \\
Saikat Roy\textsuperscript{5} \quad
Constantin Ulrich\textsuperscript{5} \quad
Klaus Maier-Hein\textsuperscript{5} \quad
Szymon Płotka\textsuperscript{6} \quad
Xiaoxi Chen\textsuperscript{7} \\
Kang Wang\textsuperscript{8} \quad
Yang Yang\textsuperscript{8} \quad
Daguang Xu\textsuperscript{9} \quad
Kai Ding\textsuperscript{10} \quad
Yucheng Tang\textsuperscript{9} \\
Alan L. Yuille\textsuperscript{1} \quad
Zongwei Zhou\textsuperscript{1,}\thanks{Correspondence to Zongwei Zhou (\href{mailto:zzhou82@jh.edu}{\textsc{zzhou82@jh.edu}})\\This is an extension of our previous manuscript, titled \textit{``ScaleMAI: Accelerating the development of trusted datasets and AI models.''}} \\[2.5mm]
\textsuperscript{1}Johns Hopkins University \quad
\textsuperscript{2}University of Bologna \quad
\textsuperscript{3}Italian Institute of Technology \\
\textsuperscript{4}University Hospital Basel \quad
\textsuperscript{5}DKFZ \quad
\textsuperscript{6}Jagiellonian University \\
\textsuperscript{7}University of Illinois Urbana–Champaign \quad
\textsuperscript{8}University of California, San Francisco \\
\textsuperscript{9}NVIDIA \quad
\textsuperscript{10}Johns Hopkins Medicine \\[1.5mm]
{\small Code, Dataset, and Models:~\href{https://github.com/MrGiovanni/ScaleMAI}{https://github.com/MrGiovanni/ScaleMAI}}
}

\begin{document}
\maketitle

\doparttoc % Tell to minitoc to generate a toc for the parts
\faketableofcontents % Run a fake tableofcontents command for the partocs

\begin{abstract}

Large, high-quality, annotated datasets are the foundation of medical AI research, but constructing even a small, moderate-quality, annotated dataset can take years of effort from multidisciplinary teams. Although active learning can prioritize what to annotate, scaling up still requires extensive manual efforts to revise the noisy annotations. We formulate this as a missing-data problem and develop \method, a framework that unifies data annotation and model development co-evolution through an Expectation–Maximization (EM) process. In this iterative process, the AI model automatically identifies and corrects the mistakes in annotations (\textbf{Expectation}), while the refined annotated data retrain the model to improve accuracy (\textbf{Maximization}). In addition to the classical EM algorithm, ScaleMAI brings human experts into the loop to review annotations that cannot be adequately addressed by either Expectation or Maximization step ($<$5\%).

As a result, \method\ progressively creates an annotated dataset of \textbf{\numofct} CT scans (\textbf{4.8$\times$} larger than the largest public dataset, PanTS \cite{li2025pants}) including \textbf{\numofmask} per-voxel annotations for benign/malignant tumors and \numofclass\ anatomical structures. \method\ iteratively trains a model that exceeds human expert performance in tumor diagnosis (\textbf{+7\%}), and outperforms models developed from smaller, moderate-quality datasets, with statistically significant gains in tumor detection (\textbf{+10\%}) and segmentation (\textbf{+14\%}) on two prestigious benchmarks.

% \textbf{Code is attached as supplementary for peer-review.} All code, datasets, and models will be publicly released.

\end{abstract}

\section{Introduction}\label{sec:intro}

\begin{figure*}[t]
    \centering
    \includegraphics[width=1\linewidth]{figure/fig_scalemai.pdf}
    \caption{\method\ reimagines the classic Expectation–Maximization (EM) algorithm~\cite{dempster1977maximum} for the problem of building large, high-quality medical datasets when expert annotations are scarce and noisy. 
    Instead of training a model on a fixed dataset and stopping there, we let the model and the dataset improve each other in a loop. At a high level, the model first ``overfits'' to the current dataset and then acts as a critic of that same dataset: wherever its predictions and the existing annotations disagree strongly, we treat this as missing or unreliable information. In the \emph{\textbf{Expectation}} step, automatic tools (Label Verifier and Label Expert) use this disagreement to correct easy annotation errors and highlight only the most doubtful regions for human review. In the \emph{\textbf{Maximization}} step, human experts focus on those few flagged cases, refine the annotations with the help of ROC-guided prioritization, and the model is retrained on this improved dataset using a mixture of unlabeled, synthetic, and selectively sampled scans. Repeating this cycle gradually turns a small, imperfect dataset into a large, expert-level resource, while keeping human effort concentrated on the $<$5\% of annotations where the AI remains uncertain.
    }
    \label{fig:scalemai}
\end{figure*}

Medical artificial intelligence (AI) is showing striking potential to detect diseases, guide treatments, and improve patient outcomes \cite{de2018clinically,national2011reduced,xia2022felix,cao2023large}. To build reliable AI models, researchers need large, high-quality, annotated datasets that capture the complexity of human anatomy and pathology \cite{li2024abdomenatlas,qu2023annotating,li2024medshapenet}. Creating such datasets, however, is painstakingly slow. Each voxel-wise annotation requires careful review by human experts and often years of coordination across institutions. As a result, annotation quantity and quality have not kept pace with the rapid advance of AI models, leaving a widening gap between what algorithms can do and the data they have to learn from. 

This gap points to a deeper issue: data annotation and model development are still treated as two independent efforts \cite{bassi2025radgpt,li2024well}. Models rely on the quality of data, but they rarely help improve it. Annotation noises persist through training, while model advances remain constrained by limited data scale. To break this cycle, we ask an important question: \emph{Can a model help build the dataset that trains it?} In this view, the model is both a learner and an editor of its own training data.

The idea is inspired by Expectation-Maximization (EM) \cite{dempster1977maximum}, a classic algorithm for refining estimates when ground truth is only partly known. We reimagine that principle for medical AI. Our approach begins with a few publicly available datasets that have incomplete annotations (listed in \figureautorefname~\ref{fig:dataset_comparison}). First, a model reviews existing annotations and flags uncertain or inconsistent regions (\textit{Expectation}). Human experts then correct only those areas, and the improved data retrain the model (\textit{Maximization}). Repeating this cycle allows both the dataset and the model to improve together, gradually converging toward expert-level annotation quality. Unlike classical EM process, we integrate selective human feedback when both \textit{E} and \textit{M} steps fail to resolve annotation noise.

We call this \textbf{\method}, a framework that connects data and model development through an iterative loop inspired by the EM process (\S\ref{sec:method}).
This design transforms model training from a passive process into a self-reinforcing process. The AI no longer consumes data blindly—it audits, questions, and learns from it. Human experts step in only when the model is uncertain, focusing on the small portion of annotations that need direct attention ($<$5\% of the dataset). This targeted workflow reduces years of work to a few months of focused refinement. The process continues until both the annotations and the model achieve a level of performance comparable to human experts.

We applied this \method\ framework to abdominal CT imaging, making two contributions.

\begin{enumerate}

    \item \textbf{An open, expert-level annotated dataset} (\S\ref{sec:dataset}), comprising \numofct\ CT scans with precise per-voxel annotations of benign and malignant pancreatic tumors, along with \numofclass\ surrounding structures. Sourced from \numofhospital\ hospitals, this dataset includes imaging metadata such as patient sex, age, contrast phase, diagnosis, spacing, and scanner details, and also includes structured and narrative radiology reports. 
    % This dataset enables a wide range of medical imaging tasks---detection, segmentation, and diagnosis---and clinical tasks such as tumor staging and radiotherapy planning.
    
    \item \textbf{An open, expert-level performing model} (\S\ref{sec:result}), developed through progressive EM process, can exceed human expert performance in tumor diagnosis (+7\%) and significantly outperforms models trained on smaller, moderate-quality datasets, achieving notable gains in detection (+10\%), and segmentation (+14\%) across two prestigious tumor benchmarks. 
    % Moreover, our \flagship\ extends its utility to tumor staging (T1--T4) and radiotherapy planning, where it can perform tumor and multi-organ segmentation on planning CT scans (distinct from the diagnostic CT scans used in training).
    
\end{enumerate}

\section{\method}\label{sec:method}

\subsection{The Expectation Step}\label{sec:method_expectation}

\subsubsection{Label Verifier}\label{sec:label_verifier}

\noindent\textbf{\textit{Intuition.}} We assume that if a segmentation model can be trained on a large dataset and then reproduce the existing annotations on held-out scans, those annotations are at least self-consistent. In contrast, when the model strongly disagrees with an annotation, it is a signal that something is missing or structurally wrong in that annotation.

\smallskip\noindent\textbf{\textit{Mechanism.}}
Label Verifier is an automatic quality controller in the \textit{Expectation step} that identifies and refines noisy annotations. We train an nnU\mbox{-}Net~\cite{isensee2021nnu} (\( \mathcal{M}_{\text{verifier}} \)) on the incompletely annotated dataset \( \mathcal{D}_{\text{\dataset-pseudo}} \) and evaluate it on the same dataset to assess annotation reliability. For each anatomical structure, we compute the Dice Similarity Coefficient (DSC) between \( \mathcal{M}_{\text{verifier}} \)’s prediction and the existing pseudo annotation in  \( \mathcal{D}_{\text{\dataset-pseudo}} \). High DSC indicates self\mbox{-}consistency with \( \mathcal{M}_{\text{verifier}} \) learning, while low DSC suggests potential noise or structural inconsistency, such as missing annotated regions or mismatches introduced by different annotation protocols. 

\smallskip\noindent\textbf{\textit{Update rule.}} To avoid over-trusting 
$\mathcal{M}_{\text{verifier}}$, we only perform automatic replacement in the extreme case \(\mathrm{DSC}=0\), where model and annotation have no spatial overlap (either one is empty and the other is non-empty, or both are in disjoint locations). In these cases, at least one of the two must be wrong, and our validation below shows the model prediction is almost always preferred by human experts. For all other DSC values, Label Verifier does not overwrite the annotation but simply forwards the case to the next stage (Label Expert or human review).

\smallskip\noindent\textbf{\textit{Validation.}}
We have Label Verifier detect and refine \(12{,}000\) noisy pseudo annotations. From a random sample of 600 cases, two human experts confirmed all 600 as true errors and preferred the replacements from \( \mathcal{M}_{\text{verifier}} \) over the original pseudo annotations. Label Verifier refined an average of 35.6\% of noisy pseudo annotations (\appendixautorefname~\tableautorefname~\ref{tab:supp_label_verifier_label_expert}).

\begin{figure}[t]
    \centering
    \includegraphics[width=1\linewidth]{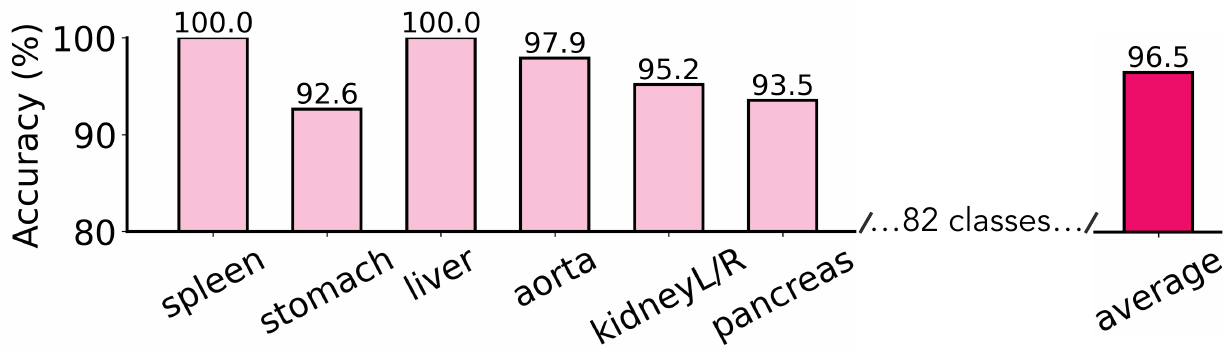}
    \caption{\textbf{Label Expert selects higher-quality annotations across diverse anatomical structures.} Evaluated on a 3,000 CT scan validation set, Label Expert achieves 96.5\% accuracy across all 88 classes. We report results on organ-at-risk for pancreatic tumors. Label Expert consistently chooses the better annotation, including challenging cases such as the pancreas, where it correctly selected 116 of 124 comparisons (93.5\% accuracy). This indicates its effectiveness in identifying higher-quality labels.}

    \label{fig:label_expert_accuracy}
\end{figure}

\subsubsection{Label Expert}\label{sec:label_expert}

\noindent\textbf{\textit{Intuition.}} A majority of annotation errors are obvious even to non-experts and can be detected by vision-language models (VLMs) \cite{wang2024qwen2} trained on diverse and extensive image-text datasets, given their strong performance in various image understanding tasks \cite{radford2021learning,jia2021scaling,alayrac2022flamingo,li2023blip,liu2024visual}. Examples of such obvious errors include organ misplacement, abnormal shapes, disconnections, multiple predictions for a single organ, noise artifacts, and annotation inconsistencies due to poor CT quality. 

\smallskip\noindent\textbf{\textit{Mechanism.}} We propose Label Expert to select the higher quality annotations using a VLM guided by anatomical knowledge. At the initialization step for EM process, Label Expert compares model predictions generated by 19 existing models from the Touchstone Benchmark \cite{bassi2024touchstone}, with the existing pseudo annotations in \( \mathcal{D}_{\text{\dataset-pseudo}} \). During the EM process loop, Label Expert compares predictions with low DSC ($<$50\%) from \( \mathcal{M}_{\text{verifier}} \) (\S\ref{sec:label_verifier}) with the existing pseudo annotations in \( \mathcal{D}_{\text{\dataset-pseudo}} \). Since pre-existing VLMs are trained on 2D natural images, they cannot directly analyze 3D CT scans. To address this, we project 3D CT scans and labels into 2D images, using a front-view projection. These projections resemble 2D X-rays with overlaid labels in red, as shown in \figureautorefname~\ref{fig:scalemai}. The VLM then evaluates these projections with prompts designed to guide its decision-making. We use the aorta as an example. The prompt teaches the VLM that `\textit{aorta should appear as a long vertical red line with a curve at the top.}' When comparing two annotations, the VLM determines that annotation \#1 matches the description better than annotation \#2.

\smallskip\noindent\textbf{\textit{Selection strategy.}} We use a lightweight tournament scheme: all pseudo annotations start from ShapeKit~\cite{liu2025shapekit} post-processing; the VLM then compare each candidate annotation sequentially and in each pairwise comparison, we keep whichever annotation the VLM prefers. After all comparisons, the remaining annotation is taken as the updated pseudo annotation.

\smallskip\noindent\textbf{\textit{Validation.}} 
We evaluate Label Expert on a held-out validation set of 3,000 CT scans, each providing expert-created per-voxel annotations for all anatomical structures present in the scan. Across \numofclass\ classes, Label Expert achieves 96.5\% accuracy in identifying the better annotation. For organ-at-risk structures relevant to pancreatic tumors (\figureautorefname~\ref{fig:label_expert_accuracy}), it also performs reliably: it selects the better pancreas annotation in 116 of 124 comparisons (93.5\%) and the better aorta annotation in 1,674 of 1,710 comparisons (97.9\%). By automating 34.7 million pairwise comparisons, Label Expert reduces human review to under 5\%, while traditional error-detection methods miss roughly 80\% of such errors~\cite{valindria2017reverse, bernard2018deep}.

\begin{figure}[t]
    \centering
    \includegraphics[width=1\linewidth]{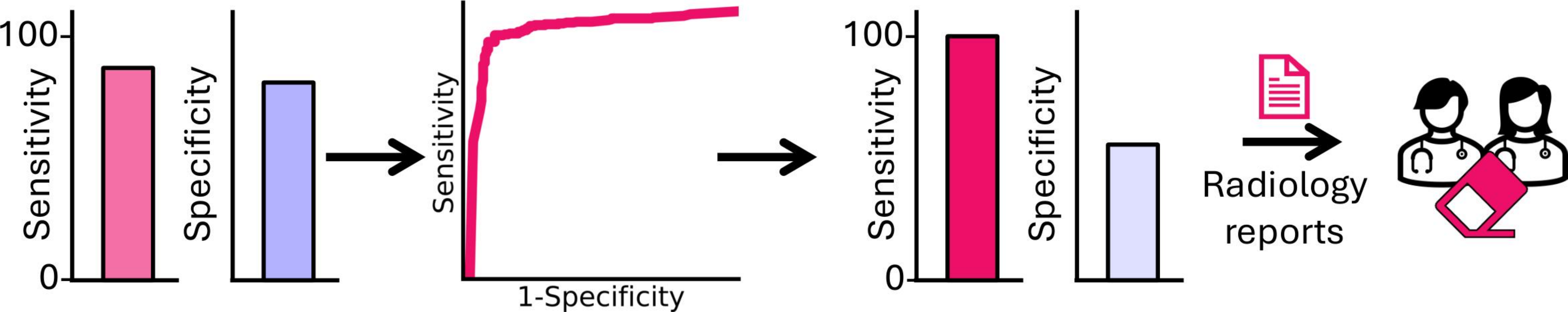}
    \caption{\textbf{ROC analysis for pancreatic tumor annotation.} Annotating per-voxel tumors is time-consuming. Our ROC analysis strategy biases AI predictions toward high sensitivity. Inevitably, this generates more false positives, but removing them is much faster and easier ($<$5 sec/tumor) than creating annotations from scratch (4--5 min/tumor). False positives in non-tumor CT scans can be automatically removed using radiology reports, and false positives in tumor CT scans can be erased with a few clicks. We achieved 99\% sensitivity for pancreatic tumor detection with only 0.6 false positives per scan---reducing annotation time by up to 92\% comparing to traditional methods.
    }
    \label{fig:roc_analysis}
\end{figure}

\subsection{The Maximization Step}\label{sec:method_maximization}

\subsubsection{ROC Analysis}\label{sec:roc_analysis}

\noindent\textbf{\textit{Intuition.}} For pancreatic tumor annotation, missing a tumor is far more costly than marking a few extra false positives. We observed that creating a new tumor mask from scratch takes 4–5 minutes, while removing an AI-generated false positive takes only a few seconds. This motivates an annotation strategy that prioritizes high sensitivity, even if it produces additional false positives that are easy to clean up.

\smallskip\noindent\textbf{\textit{Mechanism.}} We propose an efficient strategy, called \textit{ROC analysis}, to assist human experts in annotating tumors within a large-scale dataset (e.g., \numofct\ CT scans in \dataset). During the EM loop, the improving models produce increasingly accurate tumor predictions. We analyze its receiver operating characteristic (ROC) curve to select a prediction threshold that achieves near-perfect sensitivity while keeping false positives manageable, as shown in \figureautorefname~\ref{fig:roc_analysis}. (1) False positives in non-tumor CTs are automatically removed using radiology reports. (2) False positives in tumor CTs are quickly corrected using open-source tools \cite{cardoso2022monai} that allow radiologists to erase small regions with a few clicks ($<$5 seconds).

\smallskip\noindent\textbf{\textit{Validation.}} This sensitivity-first strategy dramatically reduces expert workload. The approach achieves 99\% sensitivity with 0.6 false positives per scan, reducing annotation time by up to 92\% compared to annotating every tumor voxel from scratch.

\subsubsection{Continual Tuning}\label{sec:continual-finetuning}

\noindent\textbf{\textit{Mechanism.}} To optimize the training of \flagship, we incorporate a combination of data mix and data annealing strategies. \textit{\textbf{Data mix}} consists of three primary data types. \textit{First}, unlabeled data supports self-supervised representation learning. This approach leverages the large amount of raw clinical CT scans produced daily, requiring no manual annotations. The learned representations effectively regularize \flagship, enabling faster and more efficient learning of segmentation tasks with reduced reliance on annotated data \cite{zhou2019models,zhou2021models}. \textit{Second}, synthetic data introduces variations that may not be fully represented in the training dataset, such as differences in patient demographics, scanner types, contrast phases, or tumor characteristics (e.g., location, shape, texture, size, intensity) \cite{hu2023label,hu2023synthetic,hu2022synthetic,du2024boosting}. This diversity helps the \flagship\ adapt better to out-of-distribution cases, improving its generalization. \textit{Third}, selective data targets the most challenging regions of CT scans as identified by loss function during training \cite{chou2024embracing,chen2023making,zhou2017fine,zhou2021active,zhou2019integrating}. By prioritizing repeated sampling of these regions, we can avoid \flagship\ learning from non-informative areas such as air, bedding, or irrelevant anatomical regions. This ensures that \flagship\ focuses on clinically relevant areas, such as the pancreas or abdominal region. Finally, we apply \textbf{\textit{data annealing}} \cite{dubey2024llama} by fine-tuning \flagship\ on a subset of scans with expert-created per-voxel annotations for all classes.

\smallskip\noindent\textbf{\textit{Validation.}} We evaluate \flagship\ on tumor diagnosis, detection, and segmentation as detailed in \S\ref{sec:result}.

\subsection{An Executable Summary}

At each EM loop, every annotation passes through the following decision flow:
\begin{itemize}

    \item \textit{Label Verifier:} If the model ($\mathcal{M}_{\text{verifier}}$) prediction and the current pseudo annotation have no overlap (DSC$=$0), we automatically replace the annotation with the $\mathcal{M}_{\text{verifier}}$ prediction. This captures clear structural errors.
    
    \item \textit{Label Expert:} For remaining cases with low but non-zero consistency (DSC$<$0.5), we call Label Expert and it compares the candidate annotations and selects the better one.

    \item \textit{Human escalation:} If Label Expert still cannot confidently resolve a case (e.g., conflicting VLM signals, extremely unusual anatomy, or low model confidence across all candidates), the case is escalated to human experts in the Maximization step. In practice, less than 5\% of annotations follow this path.
\end{itemize}

\noindent Thresholds (DSC$=$0 for automatic replacement and DSC$<$0.5 for VLM review) were chosen empirically to maximize precision on the held-out validation set (3,000 CT scans) and remain fixed throughout all experiments.

\section{Contribution \#1: \dataset\ Dataset}\label{sec:dataset}

\begin{figure}[t]
    \centering
    \includegraphics[width=1\linewidth]{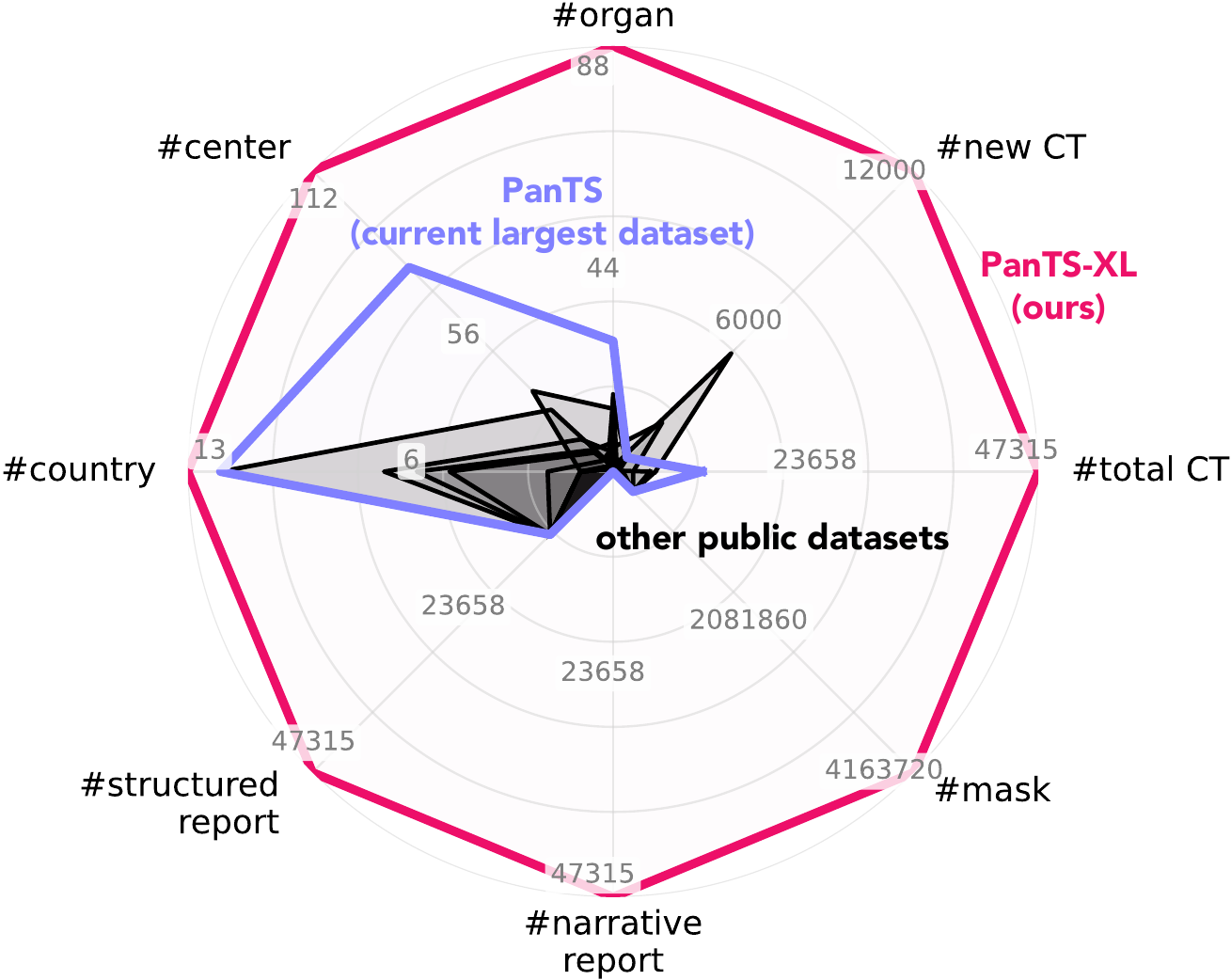}
    \caption{
    \textbf{Comparison of pancreatic and abdominal CT datasets.} We compare \dataset\ with public datasets along eight axes: number of CT scans, first-time public scans, annotated structures, contributing centers, contributing countries, availability of structured and narrative reports, and total per-voxel annotations. \textit{Earlier pancreatic and abdominal datasets} were already benchmarked in the PanTS study~\cite{li2025pants}; therefore, our comparison focuses on \textit{\textcolor{radiologist}{PanTS}} and \textit{\textcolor{flagship}{\dataset}}. A detailed comparison with public available datasets is provided in \appendixautorefname~\ref{sec:supp_related_datasets}.}
    \label{fig:dataset_comparison}
\end{figure}

\subsection{Dataset Overview}\label{sec:intro_to_dataset}

We construct \dataset, a large-scale CT dataset comprising \numofct\ scans with per-voxel annotations for pancreatic tumors and \numofclass\ surrounding anatomical structures, sourced from \numofhospital\ centers across \numofcountry\ countries. To ensure annotation consistency, we developed a rigorous standard grounded in human sectional anatomy~\citep{dixon2017human} and used it to guide human experts throughout the refinement process (\appendixautorefname~\ref{sec:supp_annotation_standard}). In addition to annotations, \dataset\ includes \numofnewct\ first-time public CT scans, paired structured and narrative radiology reports, and comprehensive imaging metadata, including scanner types, contrast phases, and patient demographics (age and sex). Figure~\ref{fig:dataset_comparison} compares \dataset\ with prior CT datasets along several axes, including number of scans, annotated structures, contributing centers and countries, availability of clinical reports, and total per-voxel annotations. Because earlier public tumor and organ datasets were already benchmarked in the PanTS study~\cite{li2025pants}, we focus our discussion on the comparison with PanTS, the previous largest one. Relative to PanTS, \dataset\ (1) contributes \numofnewct\ newly released CT scans, (2) expands the annotated classes from 27 to \numofclass, (3) increases total per-voxel annotations from 277{,}228 to 4{,}163{,}720 (15$\times$), and (4) increases source diversity from 76 to \numofhospital\ centers across \numofcountry\ countries. \textit{We will make \dataset\ publicly available.}

\begin{figure}[t]
    \centering
    \includegraphics[width=1\linewidth]{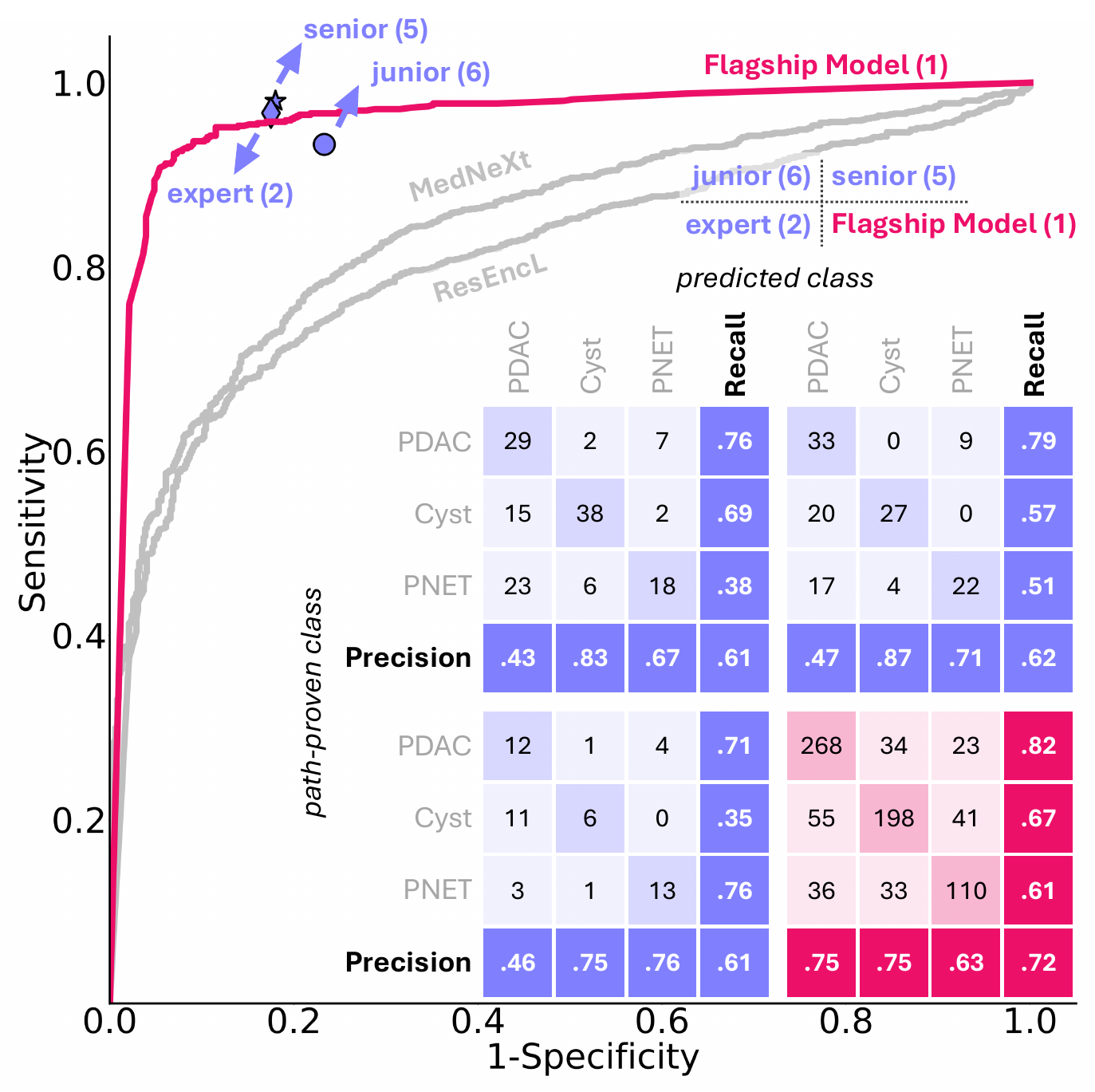}
    
    \caption{\textbf{\flagship\ matches human readers in \textit{tumor detection} and surpasses them in \textit{tumor diagnosis}.} We compare \flagship\ with 13 human readers (6 junior, 5 senior, 2 expert) on pancreatic tumor detection and diagnosis. Each reader independently evaluated 50 patients (100 contrast-enhanced CT scans); \flagship\ was evaluated on a larger cohort of 982 patients (1,964 scans). \textbf{\textit{Tumor detection.}} ROC curves (top left) show that \textcolor{flagship}{\flagship} achieves an AUC of 0.961, surpassing MedNeXt~\cite{roy2023mednext} (0.846) by 13.5\% and ResEncL~\cite{isensee2024nnu} (0.810) by 15.1\%, while matching the sensitivity–specificity performance of \textcolor{radiologist}{human readers}. \textbf{\textit{Tumor diagnosis.}} Confusion matrices (bottom right) show that \textcolor{flagship}{\flagship} attains 72\% accuracy, outperforming \textcolor{radiologist}{junior} (61\%; +11\%), \textcolor{radiologist}{senior} (66\%; +6\%), and \textcolor{radiologist}{expert readers} (69\%; +3\%) across PDAC, cyst, and PNET classification. Additional reader-study analyses are provided in \S\ref{sec:supp_reader_study}.}
    \label{fig:reader_study}
\end{figure}

\subsection{Gold Standard~vs.~Silver Standard Annotation}
\label{sec:silver_annotation_standard}

\begin{table*}[t]
\centering
\scriptsize
\caption{
\textbf{Significance of \dataset\ dataset.} AI models trained on \dataset\ significantly outperform those trained on smaller datasets in terms of generalization ability. We compare the segmentation performance of nnU-Net \cite{isensee2024nnu} trained on BTCV, WORD, AbdomenAtlas 1.0, and \dataset, evaluated on a manually annotated out-of-distribution proprietary dataset ($N$=300). The model trained on \dataset\ achieved the highest DSC scores across all anatomical structures, demonstrating superior robustness and generalization. Performance is reported as median (Q1--Q3) of DSC scores, where Q1 and Q3 denote the first and third quartiles. In addition, we have further performed a one-sided Wilcoxon signed rank test between the best-performing model and others~\cite{wiesenfarth2021methods}. The performance gain is statistically significant at the $P=0.05$ level, with highlighting in a \textcolor{flagship!75}{pink} box.
}
\begin{tabular}{p{0.14\linewidth}P{0.05\linewidth}P{0.065\linewidth}P{0.11\linewidth}P{0.11\linewidth}P{0.11\linewidth}P{0.11\linewidth}P{0.11\linewidth}}
    \toprule
      & & & \multicolumn{5}{c}{out-of-distribution test on the proprietary dataset ($N$=300)} \\
    \cmidrule(lr){4-8}
    training dataset & \# of CTs & annotators & spleen & kidneyR & kidneyL & gallbladder & liver \\
    \midrule
    BTCV~\cite{landman2015miccai} & 47 & human & 93.6\tiny{~(75.6--95.4)} & 43.9\tiny{~(0.9--89.9)} & 94.8\tiny{~(93.0--95.5)} & 77.1\tiny{~(30.5--88.3)} & 95.4\tiny{~(94.7--95.9)} \\ 
    WORD~\cite{luo2021word} & 120  & human & 93.0\tiny{~(89.7--94.3)} & 95.5\tiny{~(94.9--95.9)} & 95.1\tiny{~(92.8--95.8)} & 78.5\tiny{~(51.6--86.5)} & 94.8\tiny{~(93.8--95.5)} \\ 
    AbdomenAtlas 1.0~\cite{qu2023annotating} & 5,195 & human-AI & 95.8\tiny{~(95.1--96.5)} & 93.2\tiny{~(91.9--94.4)} & 92.8\tiny{~(91.3--93.9)} & \cellcolor{flagship!12}88.2\tiny{~(82.0--90.9)}& 96.4\tiny{~(95.8--96.9)} \\ 
    \midrule
    \dataset & \numofct\ & human-AI & 
    \cellcolor{flagship!12}\textbf{96.2\tiny{~(95.2--96.9)}} &
    \cellcolor{flagship!12}\textbf{97.7\tiny{~(97.4--98.0)}} &
    \cellcolor{flagship!12}\textbf{97.6\tiny{~(97.3--97.9)}} &
    \cellcolor{flagship!12}\textbf{88.5\tiny{~(80.6--92.1)}} &
    \cellcolor{flagship!12}\textbf{96.7\tiny{~(96.2--97.2)} } \\ 
    $\Delta$ & & & \textbf{\textcolor{red}{+0.4}} & \textbf{\textcolor{red}{+2.2}} & \textbf{\textcolor{red}{+2.5}} & \textbf{\textcolor{red}{+0.3}} & \textbf{\textcolor{red}{+0.3}} \\
    \midrule
     &  &  & stomach & aorta & postcava & pancreas & average \\
    \cmidrule(lr){4-8}
    BTCV~\cite{landman2015miccai} & 47 & human & 92.0\tiny{~(87.1--94.0)} & 61.3\tiny{~(19.6--83.3)} & 69.1\tiny{~(36.8--80.6)} & 74.5\tiny{~(66.6--79.5)} & 72.5\tiny{~(61.8--81.3)} \\ 
    WORD~\cite{luo2021word} & 120  & human & 90.7\tiny{~(87.6--92.6)} & - & - & 75.9\tiny{~(68.2--80.9)} & 87.1\tiny{~(80.8--89.8)} \\ 
    AbdomenAtlas 1.0~\cite{qu2023annotating} & 5,195 & human-AI & 94.7\tiny{~(93.0--95.5)} & 90.4\tiny{~(87.6--91.8)} & 81.2\tiny{~(75.1--84.9)} & 82.9\tiny{~(78.7--85.9)} & 89.8\tiny{~(87.9--91.2)}\\ 
    \midrule
    \dataset & \numofct\ & human-AI & 
    \cellcolor{flagship!12}\textbf{95.8\tiny{~(94.3--96.4)} } &
    \cellcolor{flagship!12}\textbf{91.8\tiny{~(88.2--94.4)} } &
    \cellcolor{flagship!12}\textbf{85.8\tiny{~(82.1--88.8)} } &
    \cellcolor{flagship!12}\textbf{85.7\tiny{~(81.8--88.1)} } &
    \cellcolor{flagship!12}\textbf{92.0\tiny{~(89.7--93.2)}} \\
    $\Delta$ & & & \textbf{\textcolor{red}{+1.1}} & \textbf{\textcolor{red}{+1.4}} & \textbf{\textcolor{red}{+4.6}} & \textbf{\textcolor{red}{+2.8}} & \textbf{\textcolor{red}{+2.2}} \\
    \bottomrule
\end{tabular}
\label{tab:high_quality_anatomical_structure_annotation}
\end{table*}

We distinguish two annotation types.
\textit{(1) Gold standard annotations} are created by human experts and verified by pathology, providing highly accurate labels but available only in limited quantity due to the clinical verification workflow. \textit{(2) Silver standard annotations} are created using imaging alone, without pathology confirmation, enabling far larger datasets. Gold standard annotations are essential for assessing clinical correctness, while silver standard datasets increase the scale and diversity needed to train high-performing models (\tableautorefname~\ref{tab:combined_detection_segmentation}). \dataset\ and \flagship\ improve together through \method. As demonstrated in the reader study (\S\ref{sec:reader_study}) and anatomical structure evaluation (\S\ref{sec:high_quality_anatomy}), \flagship\ eventually reaches human-expert–level performance in tumor detection and diagnosis (\figureautorefname~\ref{fig:reader_study}), and produces anatomical structure annotations whose quality is validated on an independent, manually annotated test set (\tableautorefname~\ref{tab:high_quality_anatomical_structure_annotation}). Once \flagship\ achieves this level of accuracy, further human refinement provides little additional benefit. We therefore stop \method\ and output \( \mathcal{D}_{\text{\dataset-pseudo}} \) as \dataset, treating it as a silver-standard dataset produced by a model whose performance matches that of trained human experts.

\subsubsection{Reader Study: Tumor Detection \& Diagnosis}
\label{sec:reader_study}

We conducted a multi-institution, multi-reader study to compare \flagship\ with human readers of varying experience levels on \textit{tumor detection} and \textit{diagnosis}.

\smallskip\noindent\textbf{\textit{Settings.}}
Thirteen board-certified human readers participated: 6 juniors ($<$8 years), 5 seniors (8--15 years), and 2 experts ($>$15 years). Each reader independently reviewed contrast-enhanced abdominal CT scans from 50 patients (100 scans), covering normal cases and three pancreatic tumor subtypes (cysts, PDAC, PNET). Readers were blinded to case distribution and performed both tumor localization (3D point marking) and subtype classification. \flagship\ was evaluated under identical settings on a larger cohort of 982 patients (1,964 scans).

\smallskip\noindent\textbf{\textit{Results.}}
As shown in Fig.~\ref{fig:reader_study}, \flagship\ achieves human-expert-level tumor detection performance, with an AUC of 0.961, providing a +13.5\% gain over MedNeXt (0.846) and +15.1\% over ResEncL (0.810). In tumor diagnosis, \flagship\ attains 72\% accuracy, surpassing junior (+11\%, 61\%), senior (+6\%, 66\%), and expert readers (+3\%, 69\%). Class-wise results (PDAC, cyst, PNET) are detailed in \S\ref{sec:tumor_diagnosis}. Additional analyses for size-stratified detection performance are provided in \appendixautorefname~\ref{sec:supp_reader_study}.

\subsubsection{High Quality Anatomical Structure Annotation} \label{sec:high_quality_anatomy}

We assess the quality of anatomical structure annotations produced through \method\ by comparing nnU-Net models trained on four abdominal CT datasets, including BTCV~\cite{landman2015miccai}, WORD~\cite{luo2022word}, AbdomenAtlas~1.0~\cite{qu2023annotating}, and our \dataset. All models were evaluated on the same manually annotated proprietary test set ($N{=}300$). We focused on nine organs-at-risk for pancreatic tumors, as these structures directly affect downstream tasks such as tumor staging and radiotherapy planning. As shown in \tableautorefname~\ref{tab:high_quality_anatomical_structure_annotation}, the model trained on \dataset\ achieves the highest DSC across \emph{all} evaluated structures, with consistent gains over models trained on smaller, manually annotated datasets. Median improvements range from +0.3\% to +4.6\%, indicating that anatomical structure annotations produced through \method\ support stronger out-of-distribution generalization.

\section{Contribution \#2: \flagship}\label{sec:result}

\noindent\textbf{\textit{Baselines}}.
We compare \flagship\ with three groups of publicly available baselines: (1) Swin~UNETR~\cite{tang2022self}, the top-performing model on the MSD leaderboard~\cite{antonelli2022medical}; (2) DTI~\cite{liu2025ai}, the top-ranking model on the public PANORAMA benchmark~\cite{alves2024panorama}; and
(3) top-performing models from the Touchstone benchmark~\cite{bassi2024touchstone}, including MedNeXt~\cite{roy2023mednext}, nnU-Net ResEncL~\cite{isensee2024nnu}, STU-Net-B~\cite{huang2023stu}, and UniSeg~\cite{ye2023uniseg}.
Swin~UNETR is implemented within the MONAI framework~\cite{cardoso2022monai}, and other models use the self-configuring nnU-Net framework~\cite{isensee2021nnu}, providing standardized training and hyperparameter selection across datasets.

\smallskip\noindent\textbf{\textit{Evaluation metrics}}.
We evaluated the performance of baselines and \flagship\ using standard metrics for tumor detection and segmentation (\tableautorefname~\ref{tab:combined_detection_segmentation}). For tumor detection, we report Sensitivity\footnote{Sensitivity is evaluated at two levels: (1) \textit{patient-wise sensitivity}, which measures whether a patient is correctly identified as having at least one tumor (\tableautorefname~\ref{tab:combined_detection_segmentation}); and (2) \textit{tumor-wise sensitivity}, which measures whether each tumor instance is correctly detected based on intersection with ground-truth annotations (\tableautorefname~\ref{tab:supp_tumor_wise_detection} in \appendixautorefname~\ref{sec:supp_tumor_detection}).}, 
Specificity, and F1-score. For tumor segmentation, we report Dice Similarity Coefficient (DSC) and Normalized Surface Distance (NSD). Detailed definitions of all metrics are in \appendixautorefname~\ref{sec:supp_evaluation_metrics}.  

\smallskip\noindent\textbf{\textit{Datasets}}. 
We benchmark baselines and \flagship\ on three datasets: MSD-Pancreas~\cite{antonelli2022medical}, PANORAMA~\cite{alves2024panorama}, and a proprietary dataset. All datasets contain per-voxel annotations for pancreatic tumors. MSD-Pancreas ($N{=}281$) is used to train public baselines, ensuring fair comparison. For out-of-distribution (OOD) evaluation, we use PANORAMA ($N{=}1{,}964$)\footnote{PANORAMA originally included 194 MSD-Pancreas cases and 80 NIH cases~\cite{roth2015deeporgan}; these are removed for a fair OOD assessment.} and the proprietary dataset ($N{=}1{,}958$). PANORAMA provides metadata including patient sex, age, scanner type, and tumor size. The proprietary dataset offers comparable metadata, with additional information on contrast phases (arterial/venous) and tumor sub-types. See \appendixautorefname~\ref{sec:supp_dataset_attributes} for details of datasets attributes.

\begin{table*}[t]
\centering
\scriptsize
\caption{
\textbf{Pancreatic tumor detection and segmentation performance.}
We benchmark pre-existing models trained on MSD-Pancreas and PANORAMA against \flagship\ on PANORAMA ($N{=}1{,}964$)
and a proprietary dataset ($N{=}1{,}958$).
For detection, we report patient-wise sensitivity, specificity (proprietary only), and F1-score (proprietary only).
For segmentation, we report median DSC and NSD with IQR.
Best results per dataset are \textbf{bolded}. In addition, we have performed a one-sided Wilcoxon signed rank test between the best-performing model and others~\cite{wiesenfarth2021methods}. The performance gain is statistically significant at the $P=0.05$ level, with highlighting in a \textcolor{flagship!75}{pink} box.
}
\begin{tabular}{
p{0.105\linewidth}
p{0.09\linewidth}
P{0.065\linewidth}
P{0.075\linewidth}
P{0.075\linewidth}
P{0.065\linewidth}
P{0.065\linewidth}
P{0.07\linewidth}
P{0.075\linewidth}
P{0.075\linewidth}
}
\toprule
&
& \multicolumn{3}{c}{PANORAMA ($N{=}1{,}964$)$^\dag$}
& \multicolumn{5}{c}{Proprietary ($N{=}1{,}958$)} \\
\cmidrule(lr){3-5} \cmidrule(lr){6-10}
method & training set
& Sens.
& DSC
& NSD
& Sens.
& Spec.
& F1
& DSC
& NSD \\
\midrule
Swin UNETR~\cite{tang2022self} & MSD-Pancreas
& 85.5\tiny~(494/578)
& 39.4\tiny~(9.5--64.3)
& 31.9\tiny~(13.3--52.2)
& 62.7\tiny~(837/1335)
& 16.7\tiny~(104/623)
& 59.1\tiny~(1674/2831)
& 11.4\tiny~(0.0--49.1)
& 11.2\tiny~(0.0--32.5)
\\
UniSeg~\cite{ye2023uniseg} & MSD-Pancreas
& 77.9\tiny~(450/578)
& 49.8\tiny~(1.1--72.1)
& 38.7\tiny~(6.5--64.2)
& 58.9\tiny~(786/1335)
& 78.5\tiny~(485/623)
& 65.9\tiny~(1572/2385)
& 12.2\tiny~(0.0--56.9)
& 9.1\tiny~(0.0--44.6)
\\
ResEncL~\cite{isensee2024nnu} & MSD-Pancreas
& 74.9\tiny~(433/578)
& 54.1\tiny~(0.0--72.2)
& 41.0\tiny~(1.8--66.4)
& 60.8\tiny~(812/1335)
& 87.0\tiny~(542/623)
& 68.0\tiny~(1624/2387)
& 22.6\tiny~(0.0--65.6)
& 13.4\tiny~(0.0--52.4)
\\
STU-Net~\cite{huang2023stu} & MSD-Pancreas
& 77.0\tiny~(445/578)
& 51.8\tiny~(0.6--73.1)
& \cellcolor{flagship!12}41.5\tiny~(4.4--67.2)
& 58.1\tiny~(775/1335)
& 84.4\tiny~(526/623)
& 66.7\tiny~(1550/2324)
& 13.5\tiny~(0.0--62.3)
& 11.2\tiny~(0.0--48.4)
\\
MedNeXt~\cite{roy2023mednext} & MSD-Pancreas
& 73.5\tiny~(425/578)
& 54.4\tiny~(0.0--74.8)
& 40.7\tiny~(0.4--66.9)
& 63.7\tiny~(850/1335)
& 83.1\tiny~(518/623)
& 69.9\tiny~(1700/2431)
& 30.6\tiny~(0.0--69.9)
& 20.4\tiny~(0.0--60.1)
\\
DTI~\cite{liu2025ai} & PANORAMA
& --
& --
& --
& 69.6\tiny~(929/1335)
& \cellcolor{flagship!12}88.1\tiny~(549/623)
& 79.6\tiny~(1935/2431)
& 42.7\tiny~(0.0--72.1)
& 35.7\tiny~(0.0--63.2)

\\
\midrule
\flagship & \dataset
& \cellcolor{flagship!12}\textbf{88.1\tiny~(509/578)}
& \cellcolor{flagship!12}\textbf{56.8\tiny~(23.7--75.8)}
& \cellcolor{flagship!12}\textbf{44.5\tiny~(20.5--66.0)}
& \cellcolor{flagship!12}\textbf{86.2\tiny~(1151/1335)}
& \cellcolor{flagship!12}\textbf{88.3\tiny~(550/623)}
& \cellcolor{flagship!12}\textbf{84.9\tiny~(2302/2713)}
& \cellcolor{flagship!12}\textbf{68.6\tiny~(34.7--82.9)}
& \cellcolor{flagship!12}\textbf{63.3\tiny~(33.2--81.9)}
\\
$\Delta$ &
& \textbf{\textcolor{red}{+2.6}}
& \textbf{\textcolor{red}{+2.4}}
& \textbf{\textcolor{red}{+3.0}}
& \textbf{\textcolor{red}{+16.6}}
& \textbf{\textcolor{red}{+0.2}}
& \textbf{\textcolor{red}{+5.3}}
& \textbf{\textcolor{red}{+25.9}}
& \textbf{\textcolor{red}{+27.6}}
\\
\bottomrule
\end{tabular}
\begin{tablenotes}
    \item $^\dag$PANORAMA annotates only PDAC, treating all other types of pancreatic tumors and healthy pancreases as \textit{Normal}---specificity and F1-score cannot be computed.
\end{tablenotes}
\label{tab:combined_detection_segmentation}
\end{table*}

\subsection{Tumor Diagnosis (+7\% Accuracy)}\label{sec:tumor_diagnosis}

We evaluate \flagship\ on three-class tumor diagnosis (PDAC, cyst, PNET)\footnote{Due to the absence of publicly available datasets suitable for benchmarking \flagship\ diagnosis performance, our evaluation compares \flagship\ exclusively with human reader performance.}. As shown in \figureautorefname~\ref{fig:reader_study}, \flagship\ achieves an accuracy of 72\%---an average +7\% improvement over human readers, outperforming junior (61\%; +11\%), senior (66\%; +6\%), and expert readers (69\%; +3\%). For PDAC, \flagship\ attains a recall of 82\%, exceeding human readers by 6\mbox{-}11\%. For cysts, it reaches 67\% recall, improving on expert readers (57\%) by +10\%. For PNET, where human recall ranges from 38\% (junior) to 51\% (senior/expert), \flagship\ achieves 61\%, corresponding to +23\% over juniors and +10\% over senior/expert readers. Across all tumor types, \flagship\ consistently outperforms human readers, with the largest gains on PNET, the most challenging class.

\subsection{Tumor Detection (+10\% Sensitivity)}
\label{sec:tumor_detection}

We evaluate pancreatic tumor detection on PANORAMA and a proprietary dataset (\tableautorefname~\ref{tab:combined_detection_segmentation}). Across both OOD benchmarks, \flagship\ attains the highest sensitivity, improving by +2.6\% on PANORAMA and +16.6\% on the proprietary dataset---an average gain of +10\%. PANORAMA annotates only PDAC and treats all other tumors and healthy pancreases as \textit{Normal}; therefore, only sensitivity is reported. On PANORAMA, \flagship\ attains 88.1\% sensitivity, surpassing all MSD-trained
models by a +2.6\% improvement over the strongest model (Swin~UNETR: 85.5\%). Since DTI~\cite{liu2025ai} was trained on PANORAMA itself, its performance cannot be evaluated fairly on this dataset and is therefore omitted. On the larger and more diverse proprietary benchmark,
\flagship\ outperforms both MSD-trained and PANORAMA-trained models. \flagship\ achieves 86.2\% sensitivity, improving over the strongest
MSD-trained models (MedNeXt: 63.7\%) by +22.5\%, and PANORAMA-trained model (DTI: 69.6\%) by +16.6\%.
\flagship\ also achieves the highest specificity (88.3\%; +0.2\%) and F1-score (84.9\%; +5.3\%) over the strongest model on this dataset.

\begin{figure}[t]
    \centering
    \includegraphics[width=1\linewidth]{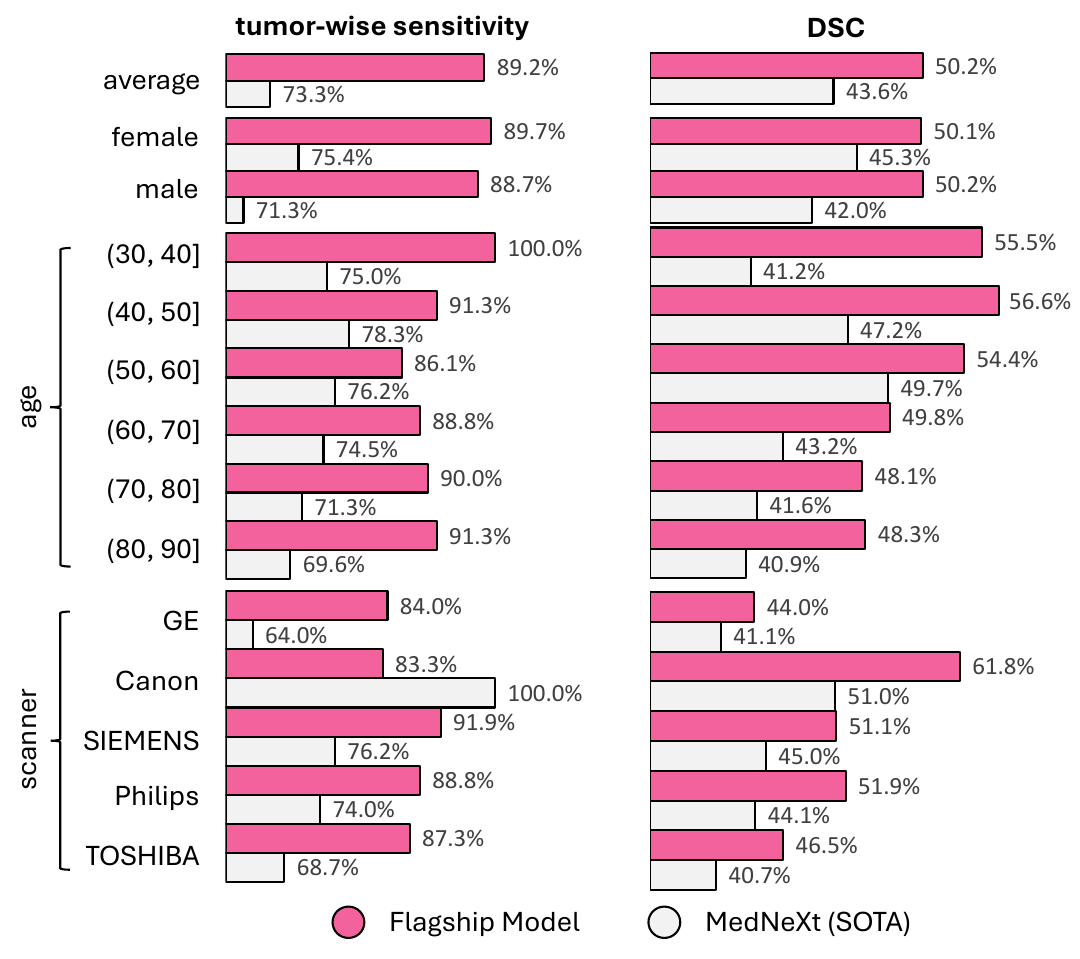}
    \caption{\textbf{\flagship\ obtains consistent gains across patient demographic and acquisition variables.} Across all reported groups on PANORAMA, \flagship\ improves over the top-performing model (MedNeXt) with average gains of +17.1\% (age), +31.7\% (sex), and +10.5\% (scanner) in sensitivity, and +8.2\% (age), +6.5\% (sex), and +6.7\% (scanner) in DSC. Results on the proprietary dataset are in \appendixautorefname~\figureautorefname~\ref{fig:supp_tumor_detection_metadata_analysis}.}  \label{fig:tumor_detection_metadata_analysis}
\end{figure}

\smallskip\noindent\textbf{\textit{Metadata analysis.}}
We analyze tumor-wise detection performance across patient demographics and acquisition variables on PANOROMA and the proprietary dataset. As shown in \figureautorefname~\ref{fig:tumor_detection_metadata_analysis}, on PANORAMA, \flagship\ yields higher sensitivity than the top-performing model (MedNeXt) across almost all reported groups, with average gains of +17.1\% across age, +31.7\% across sex, and +10.5\% across scanner manufacturers. On the proprietary dataset (see \appendixautorefname~\figureautorefname~\ref{fig:supp_tumor_detection_metadata_analysis}), \flagship\ maintains its advantage under a broader set of variables. Improvements include +23.3\% across age groups, +20.9\% across sexes, +10.5\% across races, +18.3\% across tumor sub-types, +19.3\% across tumor-sizes, and +19.3\% across contrast phases. These results demonstrate that \flagship\ preserves strong detection performance even under substantial demographic and acquisition variability.

\subsection{Tumor Segmentation (+14\% DSC)}
\label{sec:tumor_segmentation}

We evaluate tumor segmentation on PANORAMA and a proprietary dataset (\tableautorefname~\ref{tab:combined_detection_segmentation}). \flagship\ achieves the highest segmentation performance, improving DSC by +2.4\% on PANORAMA and +25.9\% on the proprietary dataset---an average gain of +14\%. On PANORAMA benchmark, \flagship\ achieves the top segmentation performance, increasing DSC from 54.4\% (MedNeXt) to 56.8\% (+2.4) and NSD from 41.5\% (STU-Net) to 44.5\% (+3.0). On the proprietary one, \flagship\ outperforms both MSD-trained and PANORAMA-trained models. It reaches 68.6\% DSC, improving over the strongest MSD-trained model (MedNeXt: 30.6\%) by +38.0 and over DTI (42.7\%) by +25.9. NSD shows a similar pattern: \flagship\ attains 63.3\%, exceeding MedNeXt (20.4\%) by +42.9 and DTI (35.7\%) by +27.6.

\smallskip\noindent\textbf{\textit{Metadata analysis.}} On PANORAMA (\figureautorefname~\ref{fig:tumor_detection_metadata_analysis}), \flagship\ achieves higher DSC than MedNeXt across all reported groups, with gains of +8.2\% (age), +6.5\% (sex), and +6.7\% (scanner). On the proprietary dataset (\appendixautorefname~\figureautorefname~\ref{fig:supp_tumor_detection_metadata_analysis}), \flagship\ again leads across all groups, with improvements of +27.9\% (age), +22.9\% (sex), +27.2\% (race), +21.3\% (tumor subtype), +23.8\% (tumor size), and +21.3\% (contrast phase). Overall, \flagship\ maintains strong segmentation performance across diverse demographic and acquisition variables.

\section{Related Work}\label{sec:related_work}

Previous studies treated data and model development separately. In contrast, we explore \textit{whether a model help build the dataset that trains it}. \method\ connects data and model improvement through an iterative process that automatically refines annotations, greatly reducing expert workload and enabling scalable medical AI.

\smallskip\noindent\textbf{\textit{Annotations.}}
Recent large-scale natural image datasets~\cite{kirillov2023segment,gadre2023datacomp,grauman2022ego4d,schuhmann2022laion} were annotated by crowdsourcing, weak supervision, or prompt-based interactive labeling. It relied on large pools of annotators and simplified 2D primitives. However, such strategies do not transfer to 3D medical imaging, especially annotating tumors, where voxel-wise annotation demands anatomical accuracy, cross-slice consistency, and domain-specific knowledge. Given these challenges, recent medical datasets speed up manual annotations through Active Learning~\cite{settles2009active,roy2018deep,beluch2018power,kasai2021reliability} and Human-in-the-Loop~\cite{holzinger2016interactive,li2021human,ma2023human}. These approaches improve annotation efficiency by prioritizing uncertain samples for expert refine. In contrast, our EM process make $>$95\% expert refinement automated to further reduce the manual efforts.

\smallskip\noindent\textbf{\textit{Datasets.}}
Earlier pancreatic CT datasets~\cite{ji2022amos,ma2024automatic} have advanced cancer-related medical image segmentation but remain limited in scale, diversity, and clinical completeness.
MSD-Pancreas~\cite{antonelli2022medical} includes only tumor (positive) cases, TCIA-Pancreas~\cite{roth2015deeporgan} includes only normal controls (negative), and PANORAMA~\cite{alves2024panorama} lacks truly normal controls because its “normal” group excludes only one tumor type and may still contain others. PanTS~\cite{li2025pants} addresses some of these gaps by aggregating 9,901 publicly available CT scans with 28 annotated structures and detailed metadata, establishing an important foundation for pancreatic tumor segmentation.
Building on this progress, our \dataset\ contributes \numofnewct\ newly released CT scans from \numofhospital\ hospitals across \numofcountry\ countries, paired structured and narrative radiology reports, and standardized clinical metadata. As a result, it supports a broad range of clinical tasks beyond segmentation—including detection, diagnosis, and report-grounded multimodal learning—and sets a new benchmark for comprehensive, scalable medical datasets.

\smallskip\noindent\textbf{\textit{Models.}} 
Recent progress in tumor segmentation has been driven by models~\cite{isensee2021nnu,isensee2024nnu,yu2020c2fnas,he2021dints,tang2022self,roy2023mednext} that achieved strong results on the open challenge and benchmark~\cite{antonelli2022medical,bassi2024touchstone} by architecture design and training strategies.
However, their performance remains limited because they depend on static datasets, where annotation quality and scale do not improve over time. Incomplete annotations and the lack of data diversity often prevent further progress, even with more advanced architectures \cite{liu2023clip,liu2024universal}.
We address this limitation by introducing an iterative EM process that continually refines both data (\dataset) and model (\flagship) quality. The model retrains itself using improved annotations, gradually approaching expert-level accuracy.

\section{Conclusion}
\label{sec:conclusion}

\method\ is directly motivated by the structure of the EM algorithm. In classical EM, the data are incomplete and the true values are treated as latent variables to be estimated. We view large-scale medical datasets in the same way: the \textit{missing data} are the unannotated or incorrectly annotated parts of each CT. In the \textit{Expectation} step, \method\ uses AI tools---Label Verifier and Expert---to propose improved annotations, filling in these missing or unreliable parts. In the \textit{Maximization} step, the updated annotations are treated as completed data to retrain the model. Repeating this process improves both the model and the annotations, mirroring the iterative refinement and self-consistency behavior of EM. Unlike classical EM, \method\ extends the E-step with VLM-based selection and selective human review, enabling data annotation at scale on real-world medical images, reducing manual efforts from years to months. \method\ produces \textbf{\dataset}, a large-scale dataset of \numofct\ CT scans---4.8$\times$ larger than the prior largest one, PanTS~\cite{li2025pants}---with \numofmask\ per-voxel annotations for \numofclass\ classes across \numofhospital\ centers. The resulting \textbf{\flagship} achieves best performance on two benchmarks, with significant gains of +7/+10/+14\% in tumor diagnosis, detection, and segmentation, respectively. 

\clearpage
% \smallskip
\noindent\textbf{Acknowledgments.}
This work was supported by the Lustgarten Foundation for Pancreatic Cancer Research and the National Institutes of Health (NIH) under Award Number R01EB037669. We would like to thank the Johns Hopkins Research IT team in \href{https://researchit.jhu.edu/}{IT@JH} for their support and infrastructure resources where some of these analyses were conducted; especially \href{https://researchit.jhu.edu/research-hpc/}{DISCOVERY HPC}. We thank Jaimie Patterson for writing a \href{}{news article} about this project. Paper content is covered by patents pending.

{
    \small
    \bibliographystyle{ieeenat_fullname}
    \bibliography{refs,zzhou}
}

\clearpage
\appendix
\setcounter{page}{1}
% \maketitlesupplementary
\onecolumn
\renewcommand \thepart{}
\renewcommand \partname{}

\part{Appendix} % Start the appendix part
\setcounter{secnumdepth}{4}
\setcounter{tocdepth}{4}
\parttoc % Insert the appendix TOC

\clearpage
\section{Technical Details of \method}\label{sec:supp_method}

\subsection{The Expectation Step}\label{sec:supp_e_step}

\subsubsection{Label Verifier}
\label{sec:supp_label_verifier}

Label Verifier serves as an automatic quality controller within the Expectation step of \method. Its role is to identify annotations that are inconsistent with the statistical patterns learned by a segmentation model $ \mathcal{M}_{\text{verifier}} $ trained on the current pseudo-annotations dataset $ \mathcal{D}_{\text{\dataset-pseudo}} $. The key intuition is that if $ \mathcal{M}_{\text{verifier}} $ can accurately reproduce an annotation on held-out scans, that annotation is likely self-consistent; conversely, strong disagreement indicates potential noise or structural mistakes. By comparing $ \mathcal{M}_{\text{verifier}} $ predictions with existing pseudo annotations for each structure, Label Verifier provides a principled mechanism to flag and refine erroneous annotations before they propagate into the next iteration of \method.

\begin{algorithm}[h]
\caption{\textbf{Label Verifier: Model-Guided Detection \& Refinement of Noisy Annotations}}
\label{alg:label_verifier}
\begin{algorithmic}[1]

\STATE \textbf{Input:} Pseudo-annotated dataset $ \mathcal{D}_{\text{\dataset-pseudo}} $; 
trained verifier model $ \mathcal{M}_{\text{verifier}} $ (nnU-Net trained on $ \mathcal{D}_{\text{\dataset-pseudo}} $); 
anatomical structure set $ \mathcal{S} $.

\STATE \textbf{Output:} Updated pseudo-annotations $ \hat{\mathcal{D}}_{\text{\dataset-pseudo}} $ (with optional replacements when DSC = 0)

\vspace{0.5em}
\STATE // Step 1: Compute agreement between model and annotation
\FOR{each sample $(x, y_{\text{pseudo}})$ in $ \mathcal{D}_{\text{\dataset-pseudo}} $}
    \STATE Predict verifier mask: $y_{\text{verifier}} \gets \mathcal{M}_{\text{verifier}}(x)$
    \FOR{each structure $s \in \mathcal{S}$}
        \STATE Extract structure-specific masks:
        \STATE \hspace{1.5em} $y_{\text{pseudo}}^s \gets$ mask of $s$ in $y_{\text{pseudo}}$
        \STATE \hspace{1.5em} $y_{\text{verifier}}^s \gets$ mask of $s$ in $y_{\text{verifier}}$
        \STATE Compute Dice score:
        \STATE \hspace{1.5em} $\mathrm{DSC}_s = \mathrm{DSC}(y_{\text{pseudo}}^s,\, y_{\text{verifier}}^s)$
    \ENDFOR

    \vspace{0.4em}
    \STATE // Step 2: Apply update rule (only in extreme disagreement)
    \FOR{each structure $s \in \mathcal{S}$}
        \IF{$\mathrm{DSC}_s = 0$}
            \STATE $y_{\text{pseudo}}^s \gets y_{\text{verifier}}^s$
            \COMMENT{Replace annotation only when $ \mathcal{M}_{\text{verifier}} $ prediction and pseudo annotation have no overlap}
        \ENDIF
    \ENDFOR

    \STATE Store updated annotation $y_{\text{pseudo}}$ in $\hat{\mathcal{D}}_{\text{\dataset-pseudo}}$
\ENDFOR

\vspace{0.4em}
\STATE \textbf{return} $\hat{\mathcal{D}}_{\text{\dataset-pseudo}}$

\end{algorithmic}
\end{algorithm}

\noindent Label Verifier quantifies the agreement between the predictions of $\mathcal{M}_{\text{verifier}}$ and the existing annotations, and selectively replaces an annotation only when there is a complete mismatch ($\mathrm{DSC} = 0$). To assess its practical impact, we measure how often Label Verifier detects and refines erroneous annotations during \method’s annotation of \dataset. Table~\ref{tab:supp_label_verifier_label_expert} summarizes representative anatomical structures with notable refinement rates, highlighting the effectiveness of Label Verifier—particularly for structures that are often absent in abdominal CT scans.

\begin{table}[h]
\centering
\scriptsize
\caption{\textbf{Label Verifier detects and refines 35.6\% of annotation errors.} Label Verifier replaces a pseudo annotation only when it has no spatial overlap with $ \mathcal{M}_{\text{verifier}} $ prediction (DSC = 0).
In total, 51,454 erroneous annotations were refined iteratively through \method. We report representative anatomical structures that show notable refinement rates. Label Verifier is particularly effective for some structures that are often absent in abdominal CT scans---such as the lung, prostate, rectum, and bladder---where existing annotations may contain false positives, resulting in no overlap with $ \mathcal{M}_{\text{verifier}} $ prediction.}
\begin{tabular}{p{0.08\linewidth}P{0.03\linewidth}P{0.03\linewidth}P{0.05\linewidth}P{0.03\linewidth}P{0.04\linewidth}P{0.03\linewidth}P{0.05\linewidth}P{0.03\linewidth}P{0.05\linewidth}P{0.03\linewidth}P{0.05\linewidth}P{0.02\linewidth}P{0.03\linewidth}P{0.03\linewidth}P{0.03\linewidth}}
\toprule
method & spleen & kidney & gallbladder & liver & stomach & aorta & pancreas & prostate & duodenum & femur & esophagus & lung & bladder & rectum & average \\
\midrule
Label Verifier & 28.0 & 24.6 & 48.8 & 8.3 & 38.4 & 10.0 & 14.0 & 52.8 & 40.8 & 48.7 & 24.6 & 47.4 & 49.6 & 49.4 & 35.6 \\
\bottomrule
\end{tabular}
\label{tab:supp_label_verifier_label_expert}
\end{table}

\clearpage

\subsubsection{Label Expert}
\label{sec:supp_label_expert}

Label Expert is designed to improve annotation quality by leveraging a VLM (e.g., Qwen2-VL~\cite{wang2024qwen2}) with detailed anatomical knowledge as prompt. Given two candidate annotations for the same CT scan---one produced during initialization or by $\mathcal{M}_{\text{verifier}}$, and the other drawn from the existing pseudo-annotation set $\mathcal{D}_{\text{\dataset-pseudo}}$ ---Label Expert determines which annotation more faithfully reflects the underlying anatomy. Because existing VLMs operate on 2D natural images, rather than volumetric medical data, each 3D CT scan is first projected into a front-view image with the corresponding annotation overlaid in red. The VLM then evaluates these projections using structure-specific prompts that define the expected anatomical appearance, location, and continuity of each organ or structure, enabling an informed comparison between two annotations.

\begin{algorithm}[h]
\caption{\textbf{Label Expert: VLM-Guided Selection of Higher-Quality Annotation}}
\label{alg:label_expert_prompt}
\begin{algorithmic}[1]

\STATE \textbf{Input:} 3D CT scan $x$; pseudo-annotation $y_{\text{pseudo}} \in \mathcal{D}_{\text{\dataset-pseudo}}$;
candidate annotation $y_{\text{cand}}$ (prediction from initialization or low-DSC output from
$\mathcal{M}_{\text{verifier}}$ during \method);
anatomical structure set $\mathcal{S} = \{\text{Aorta, Postcava, Kidneys, Liver, Pancreas, Spleen, ... (88 classes)}\}$;
VLM with structure-specific anatomical prompts.

\STATE \textbf{Output:} Selected annotation $\hat{y} \in \{y_{\text{cand}}, y_{\text{pseudo}}\}$

\vspace{0.5em}
\STATE // Step 1: Prepare the two annotations for comparison
\STATE $Annotation_1 \gets y_{\text{cand}}$
\STATE $Annotation_2 \gets y_{\text{pseudo}}$

\vspace{0.5em}
\STATE // Step 2: Generate 2D projections for VLM evaluation
\STATE $I_x \gets$ front-view projection of CT scan $x$
\STATE $I_1 \gets$ overlay $Annotation_1$ (red) on $I_x$
\STATE $I_2 \gets$ overlay $Annotation_2$ (red) on $I_x$

\vspace{0.5em}
\STATE // Step 3: Structure-wise comparison using VLM prompts
\STATE $score_1 \gets 0$, $score_2 \gets 0$

\FOR{each $structure \in \mathcal{S}$}
    \STATE Construct anatomical prompt $p(structure)$
    \STATE \hspace{1.5em} // e.g., for the aorta: “the aorta appears as a long vertical red tube with a curve at the top”
    \STATE Query VLM with $(I_1, I_2, p(structure))$ to obtain preference 
           $r_{structure} \in \{1, 2, \text{tie}\}$
    \IF{$r_{structure} = 1$}
        \STATE $score_1 \gets score_1 + 1$
    \ELSIF{$r_{structure} = 2$}
        \STATE $score_2 \gets score_2 + 1$
    \ENDIF
\ENDFOR

\vspace{0.5em}
\STATE // Step 4: Final selection
\IF{$score_1 > score_2$}
    \STATE $\hat{y} \gets Annotation_1$  \COMMENT{VLM prefers candidate annotation}
\ELSE
    \STATE $\hat{y} \gets Annotation_2$  \COMMENT{VLM prefers pseudo-annotation or tie}
\ENDIF

\STATE \textbf{return} $\hat{y}$

\end{algorithmic}
\end{algorithm}

Label Expert selects the higher-quality annotation by scoring two candidate overlays across all anatomical structures with the assistance of a VLM. To carry out this procedure, each structure is assessed using a tailored prompt that describes its expected shape, location, and anatomical context. These prompts guide the VLM’s reasoning when comparing the two projected annotations. The following prompt templates provide representative examples of the instructions given to the VLM for different structures, ensuring consistent and anatomically informed evaluation during the comparison process.

\clearpage

\begin{tcolorbox}[
  breakable,
  colframe=flagship!60,
  colback=flagship!2,
  coltitle=white,
  fonttitle=\bfseries,
  title=Label Expert Prompt: Aorta Annotation Comparison,
  colbacktitle=flagship!60
]
You are given two images, Image1 and Image2, representing frontal projections of the same 3D CT scan and visually comparable to AP X-rays. 
Each image contains a red overlay representing the aorta, and the overlays differ between the two images.

Your task is to determine which image contains the more accurate aorta annotation.

\textbf{Evaluation criteria:}
\begin{enumerate}
  \item \textit{Cranial extension} —  
        The aorta normally extends into the thoracic region; the better annotation shows a longer superior reach.

  \item \textit{Lumbar alignment} —  
        If the lumbar spine is visible, the aorta overlay should descend to an anatomically plausible level.

  \item \textit{Shape and continuity} —  
        The aorta should appear as a smooth tubular structure without fragmentation or abrupt widening.
\end{enumerate}

After reviewing both images, conclude whether Image1 or Image2 contains the more anatomically accurate aorta annotation.
\end{tcolorbox}

\begin{tcolorbox}[
  breakable,
  colframe=flagship!60,
  colback=flagship!2,
  coltitle=white,
  fonttitle=\bfseries,
  title=Label Expert Prompt: Postcava Annotation Comparison,
  colbacktitle=flagship!60
]
You are given two images, Image1 and Image2, representing frontal projections of the same 3D CT scan and visually comparable to AP X-rays. 
Each image contains a red overlay representing the postcava, with different overlays shown in the two images.

Your task is to determine which overlay more accurately represents the postcava.

\textbf{Evaluation criteria:}
\begin{enumerate}
  \item \textit{Thoracic reach} —  
        The postcava should extend into the thoracic region near the upper ribs.

  \item \textit{Lumbar descent} —  
        If the lumbar spine is visible, the postcava overlay should descend into the upper lumbar region realistically.

  \item \textit{Continuity and form} —  
        The postcava should appear as a continuous, narrow tubular structure without breaks or irregular deviations.
\end{enumerate}

After evaluating both images, determine whether Image1 or Image2 provides the more anatomically correct postcava annotation.
\end{tcolorbox}

\begin{tcolorbox}[
  breakable,
  colframe=flagship!60,
  colback=flagship!2,
  coltitle=white,
  fonttitle=\bfseries,
  title=Label Expert Prompt: Spleen Annotation Comparison,
  colbacktitle=flagship!60
]
You are given two images, Image1 and Image2, representing frontal projections of the same 3D CT scan and visually comparable to AP X-rays. 
Each image contains a red overlay representing the spleen, and the overlays differ between the two images.

Your task is to determine which overlay more accurately represents the spleen.

\textbf{Anatomical guidelines:}
\begin{enumerate}
  \item \textit{Shape} —  
        The spleen typically has an oval or crescent-like outline with smooth, continuous curvature.

  \item \textit{Smoothness} —  
        A proper spleen contour should not contain irregular protrusions, abrupt angles, internal gaps,  
        or jagged edges.

  \item \textit{Continuity} —  
        The spleen is a single anatomical structure; the overlay should therefore form one continuous region.

  \item \textit{Location} —  
        The spleen lies in the upper left abdomen (right side of the image in AP orientation),  
        beneath the ribs and diaphragm and adjacent to the stomach and left kidney.
\end{enumerate}

After reviewing both images, conclude whether Image1 or Image2 contains the more anatomically accurate spleen annotation.
\end{tcolorbox}

\begin{tcolorbox}[
  breakable,
  colframe=flagship!60,
  colback=flagship!2,
  coltitle=white,
  fonttitle=\bfseries,
  title=Label Expert Prompt: Kidney Annotation Comparison,
  colbacktitle=flagship!60
]
You are given two images, Image1 and Image2, representing frontal projections of the same 3D CT scan and visually comparable to AP X-rays. 
Each image contains red overlays representing the kidneys, which differ between the two images.

Your task is to determine which image contains the more accurate kidney annotation.

\textbf{Anatomical guidelines:}
\begin{enumerate}
  \item \textit{Number of structures} —  
        Two kidneys should be present, appearing as two separate regions.

  \item \textit{Shape} —  
        Kidneys have a bean-like outline with a concave side medially and a convex side laterally.

  \item \textit{Location} —  
        Kidneys lie lateral to the spine near the lower ribs, typically at similar vertical levels.
\end{enumerate}

After comparing both images, conclude whether Image1 or Image2 contains the more anatomically accurate kidney annotation.
\end{tcolorbox}

\begin{tcolorbox}[
  breakable,
  colframe=flagship!60,
  colback=flagship!2,
  coltitle=white,
  fonttitle=\bfseries,
  title=Label Expert Prompt: Liver Annotation Comparison,
  colbacktitle=flagship!60
]
You are given two images, Image1 and Image2, representing frontal projections of the same 3D CT scan and visually comparable to AP X-rays. 
Each image contains a red overlay representing the liver. The overlays differ between the two images.

Your task is to determine which overlay more accurately represents the liver.

\textbf{Anatomical guidelines:}
\begin{enumerate}
  \item \textit{Size and shape} —  
        The liver is a large triangular or wedge-shaped organ; the overlay should reflect a broad smooth outline.

  \item \textit{Location} —  
        It lies in the upper right abdomen (left side of the image in AP view), beneath the diaphragm  
        and partially crossing the midline.

  \item \textit{Relation to rib cage} —  
        Most of the liver is covered by ribs; the overlay should correspond to this region.
\end{enumerate}

After evaluating both images, determine whether Image1 or Image2 contains the more anatomically accurate liver annotation.
\end{tcolorbox}

\begin{tcolorbox}[
  breakable,
  colframe=flagship!60,
  colback=flagship!2,
  coltitle=white,
  fonttitle=\bfseries,
  title=Label Expert Prompt: Pancreas Annotation Comparison,
  colbacktitle=flagship!60
]
You are given two images, Image1 and Image2, representing frontal projections of the same 3D CT scan and visually comparable to AP X-rays. 
Each image contains a red overlay representing the pancreas, and the overlays differ.

Your task is to determine which overlay more accurately represents the pancreas.

\textbf{Anatomical guidelines:}
\begin{enumerate}
  \item \textit{Shape} —  
        The pancreas is an elongated organ with a thicker head and a thinner tail.

  \item \textit{Position} —  
        It lies in the upper abdomen, posterior to the stomach and near the lower ribs,  
        usually spanning horizontally with mild curvature.

  \item \textit{Continuity} —  
        The pancreas should appear as one smooth, continuous structure without disconnected components.
\end{enumerate}

After comparing both images, conclude whether Image1 or Image2 contains the more anatomically accurate pancreas annotation.
\end{tcolorbox}

\begin{tcolorbox}[
  breakable,
  colframe=flagship!60,
  colback=flagship!2,
  coltitle=white,
  fonttitle=\bfseries,
  title=Label Expert Prompt: Rib Annotation Comparison (L1--L12 \& R1--R12),
  colbacktitle=flagship!60
]
You are given two images, Image1 and Image2, representing frontal projections of the same 3D CT scan and visually comparable to AP X-rays. 
Each image contains red overlays marking the ribs (Left 1--12 and Right 1--12), which differ between the two images.

Your task is to determine which overlay more accurately represents the rib anatomy.

\textbf{Anatomical guidelines:}
\begin{enumerate}
  \item \textit{Completeness} —  
        Twelve ribs should be present on each side, without missing or duplicated structures.

  \item \textit{Symmetry} —  
        Corresponding rib pairs should have similar curvature and placement.

  \item \textit{Curvature} —  
        Ribs should appear as smooth arcs extending laterally from the spine.

  \item \textit{Ordering} —  
        The pattern should follow the natural superior-to-inferior rib order.

  \item \textit{Continuity} —  
        Each rib should be a continuous curved structure.
\end{enumerate}

After comparing both images, conclude whether Image1 or Image2 contains the more anatomically accurate rib annotation.
\end{tcolorbox}

\clearpage
\subsection{The Maximization Step} 
\label{sec:supp_m_step}

\subsubsection{ROC Analysis for Tumor Annotation}\label{sec:supp_roc_analysis}

\begin{figure*}[h]
    \centering
    \includegraphics[width=1\linewidth]{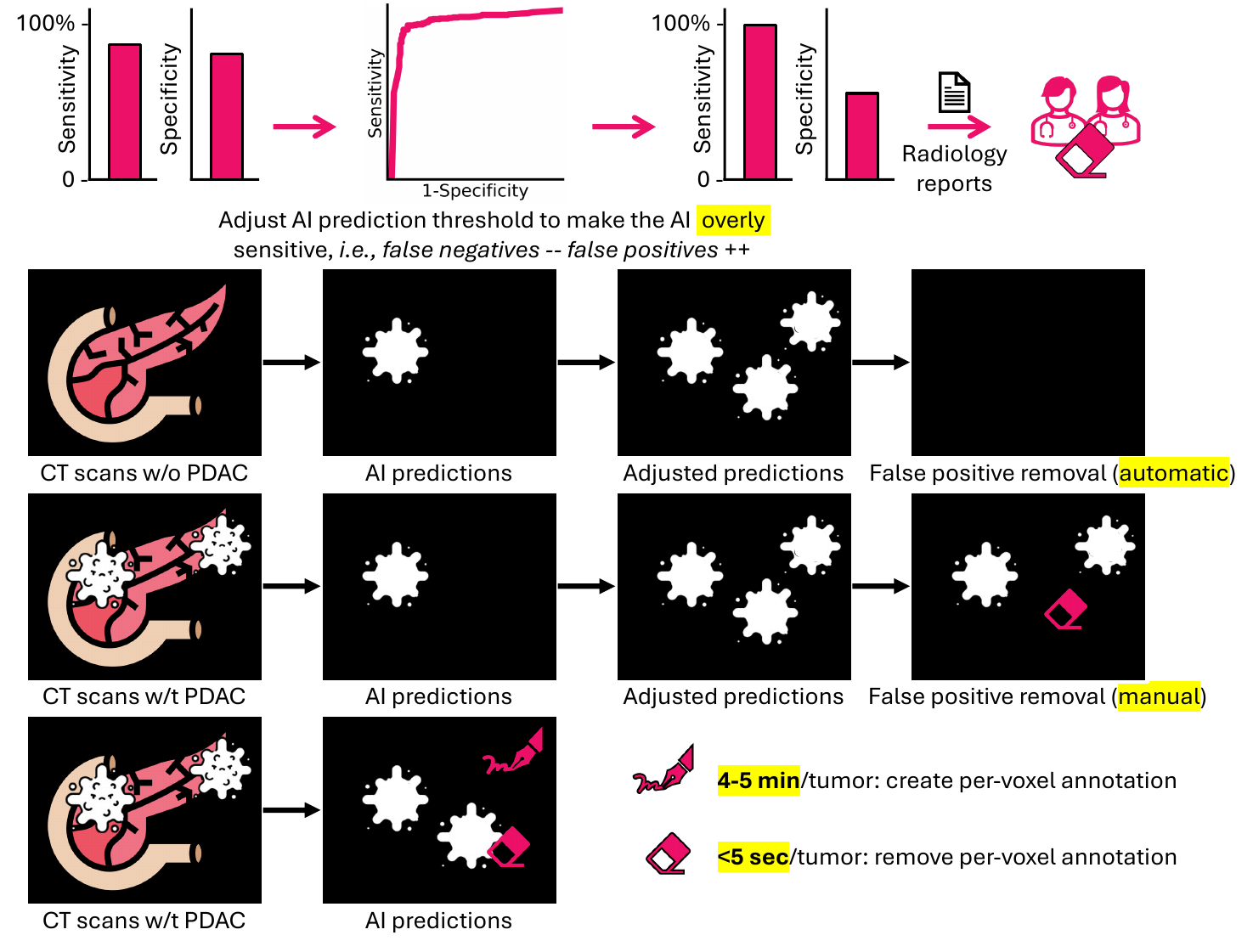}
    \caption{\textbf{ROC Analysis for Pancreatic Tumor Annotation.} We propose an efficient strategy, called ROC analysis, to assist radiologists in annotating tumors within a large-scale dataset (e.g., over 47,000 CT scans in our study). During the iterative data curation and annotation process facilitated by the \method\ framework, AI model performance improves as data quality increases. In turn, stronger AI models generate more accurate pseudo labels with high sensitivity and specificity, significantly reducing radiologists' workload. Our observations show that removing AI false positives is much faster than creating per-voxel annotations for false negatives (missed tumors). Removing a false positive takes less than five seconds, whereas creating per-voxel annotations for a missed tumor can take 4--5 minutes. This insight motivates us to analyze the AI model's receiver operating characteristic (ROC) curve, which allows us to adjust the prediction threshold to prioritize sensitivity over specificity. To minimize radiologists' workload, we aim for nearly perfect sensitivity while maintaining acceptable specificity. By intentionally biasing the model towards high sensitivity, the AI minimizes missed tumors but inevitably introduces more false positives. Since handling false positives is simpler, this trade-off optimizes efficiency: (1) \textbf{\textit{False positives in non-tumor CT scans}} can be automatically removed by cross-referencing radiology reports, which are typically available in clinical repositories (as illustrated in the second line in the Figure). (2) \textbf{\textit{False positives in tumor CT scans}} can be efficiently removed using open-source annotation tools \cite{cardoso2022monai}. These tools enable radiologists to erase false positives with a few clicks, leveraging the AI's highly sensitive per-voxel predictions. In our study, we achieved 99\% sensitivity for pancreatic tumor detection with only 0.6 false positives per scan. This means radiologists only have to remove just one false positive for every two CT scans (as illustrated in the third line in the Figure). Compared to creating per-voxel annotations from scratch, our ROC analysis approach reduces annotation time by up to 92\%, significantly streamlining the workflow.
    }
    \label{fig:supp_roc_analysis}
\end{figure*}

\clearpage
\subsubsection{Continual Tuning}\label{sec:supp_continual_tuning}

\begin{figure*}[h]
    \centering
    \includegraphics[width=1\linewidth]{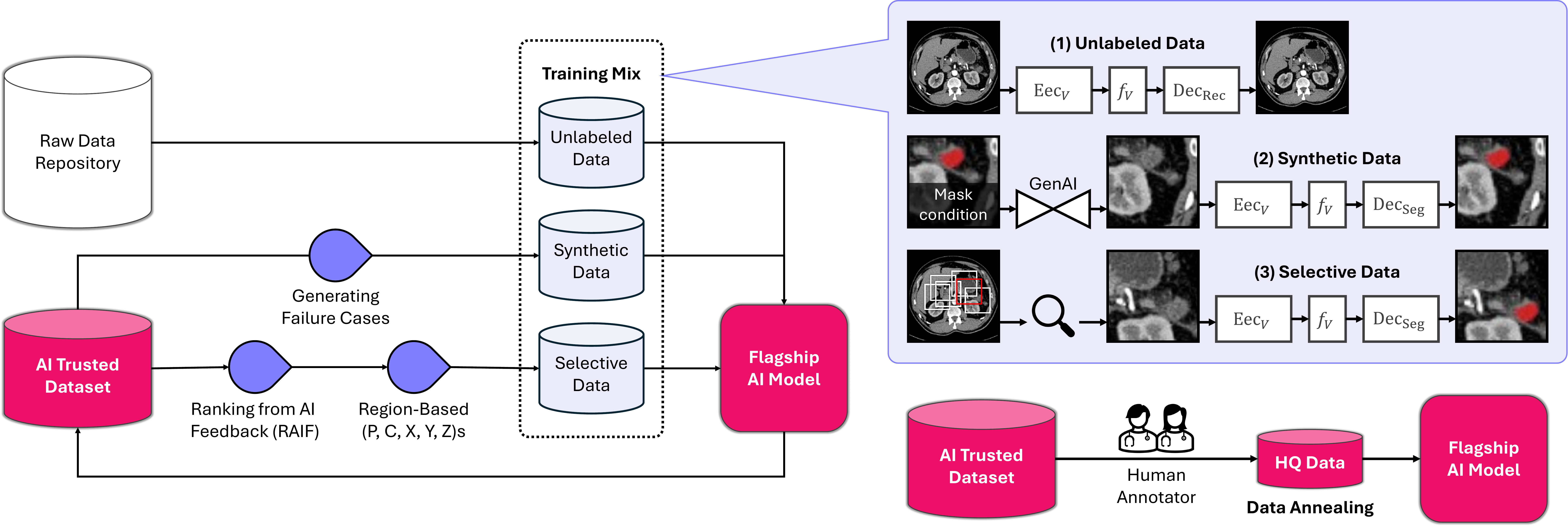}
    \caption{\textbf{Training \flagship\ with Data Mix and Data Annealing.}
    To optimize the training of \flagship, we incorporate a combination of data mix and data annealing strategies. The data mix consists of three primary types: \textit{\textbf{First}}, unlabeled data is utilized for self-supervised representation learning. This approach leverages the vast quantities of raw clinical data generated daily, requiring no manual annotation. The learned representations effectively regularize the model, enabling faster and more efficient learning of segmentation tasks with reduced reliance on annotated data. This methodology, supported by extensive literature, demonstrates the potential to exploit unlabeled clinical data for robust model training. \textit{\textbf{Second}}, synthetic data is employed to generate a diverse array of scans. These include variations across demographics, scanner types, and contrast enhancements, as well as tumors with differing locations, shapes, textures, sizes, and intensities that are not fully represented in the training set. This diversity enhances the model's robustness, particularly when encountering out-of-distribution test cases. \textit{\textbf{Third}}, selective data focuses on the most challenging regions of CT scans that confuse the model during training, as identified by the loss function. By prioritizing repeated sampling of these regions, the model learns more efficiently, avoiding the inefficiencies of processing non-informative areas such as air, bedding, or irrelevant anatomical regions. This targeted approach ensures that the model focuses on clinically relevant areas, such as the pancreas or abdominal region. \textit{\textbf{Finally}}, once the model is trained on data mix, we introduce data annealing to further fine-tune \flagship. We identify a gold-standard subset, consisting of voxel-level annotations meticulously created by expert radiologists. This data annealing technique has proven effective in large-scale training efforts in other domains, such as ChatGPT \cite{achiam2023gpt} and Llama 3 \cite{dubey2024llama}. However, in the medical field, the lack of gold-standard data and the predominance of silver-standard annotation have limited its exploration. When releasing the dataset, we will explicitly mark this gold-standard subset to facilitate further research and development in the field.
    }
    \label{fig:supp_data_mix_annealing}
\end{figure*}

\clearpage
\section{Quality Assessment of \textbf{\dataset} Datasets}\label{sec:supp_AI_trusted_dataset}

\subsection{Annotation Standard for \numofclass\ Anatomical Structures} \label{sec:supp_annotation_standard}

The annotated areas of all tubular structures include both the tube wall and the lumen but exclude surrounding tissues, such as organs, mesentery, and adipose tissue. The \ul{pancreatic duct} is identified as a low-attenuation tubular structure within the pancreas and should be marked from the tail to the ampulla of Vater. The \ul{common bile duct (CBD)} appears as a low-attenuation tubular structure and should be annotated from the confluence of the common hepatic duct and bile duct to the ampulla of Vater. The \ul{superior mesenteric artery (SMA)} is highlighted as a bright arterial structure originating from the aorta and should be traced from its origin to the point where it branches. The \ul{celiac artery} is a short vessel arising from the aorta and splitting into the left gastric, splenic, and common hepatic arteries; it should be annotated from its origin to the bifurcation. \ul{Veins} include the portal vein, splenic vein, and superior mesenteric vein: the \ul{portal vein} is traced from the confluence of the splenic and superior mesenteric veins to its intrahepatic entry; the \ul{splenic vein} is marked from the splenic hilum to the SMV confluence; and the \ul{superior mesenteric vein (SMV)} is annotated from its major tributaries to the confluence with the splenic vein. Solid organs—including the \ul{liver}, \ul{liver segments 1–8}, \ul{pancreas} (and its \ul{head}, \ul{body}, and \ul{tail}), \ul{kidneys} (left and right), \ul{adrenal glands} (left and right), \ul{spleen}, \ul{stomach}, \ul{duodenum}, \ul{intestine}, \ul{colon}, \ul{gallbladder}, \ul{esophagus}, \ul{bladder}, \ul{prostate}, and \ul{lungs} (left and right)—are annotated by including the entire parenchyma while excluding surrounding fat, adjacent organs, and extrinsic vasculature. The \ul{aorta} and \ul{postcava} are annotated as continuous tubular structures throughout their visible abdominal course, following the same lumen-plus-wall rule as other vessels. For osseous structures, each \ul{rib} (left and right, 1–12) is annotated to include the full cortical and cancellous bone while excluding costal cartilage; likewise, each vertebra—from \ul{C1–C7}, \ul{T1–T12}, and \ul{L1–L5}—is annotated to include the vertebral body, pedicles, transverse processes, and spinous process while excluding intervertebral discs and surrounding soft tissues. The \ul{femur left} and \ul{femur right} are annotated by including the entire visible femoral head, neck, and proximal shaft, capturing both cortical and cancellous bone while excluding surrounding muscle and soft tissue. These standards ensure consistent and anatomically faithful annotations across all \numofclass\ structures in the dataset.

\clearpage
\subsection{Reader Study: Tumor Detection \& Diagnosis}
\label{sec:supp_reader_study}

\begin{figure}[h]
    \centering
    \includegraphics[width=1\linewidth]{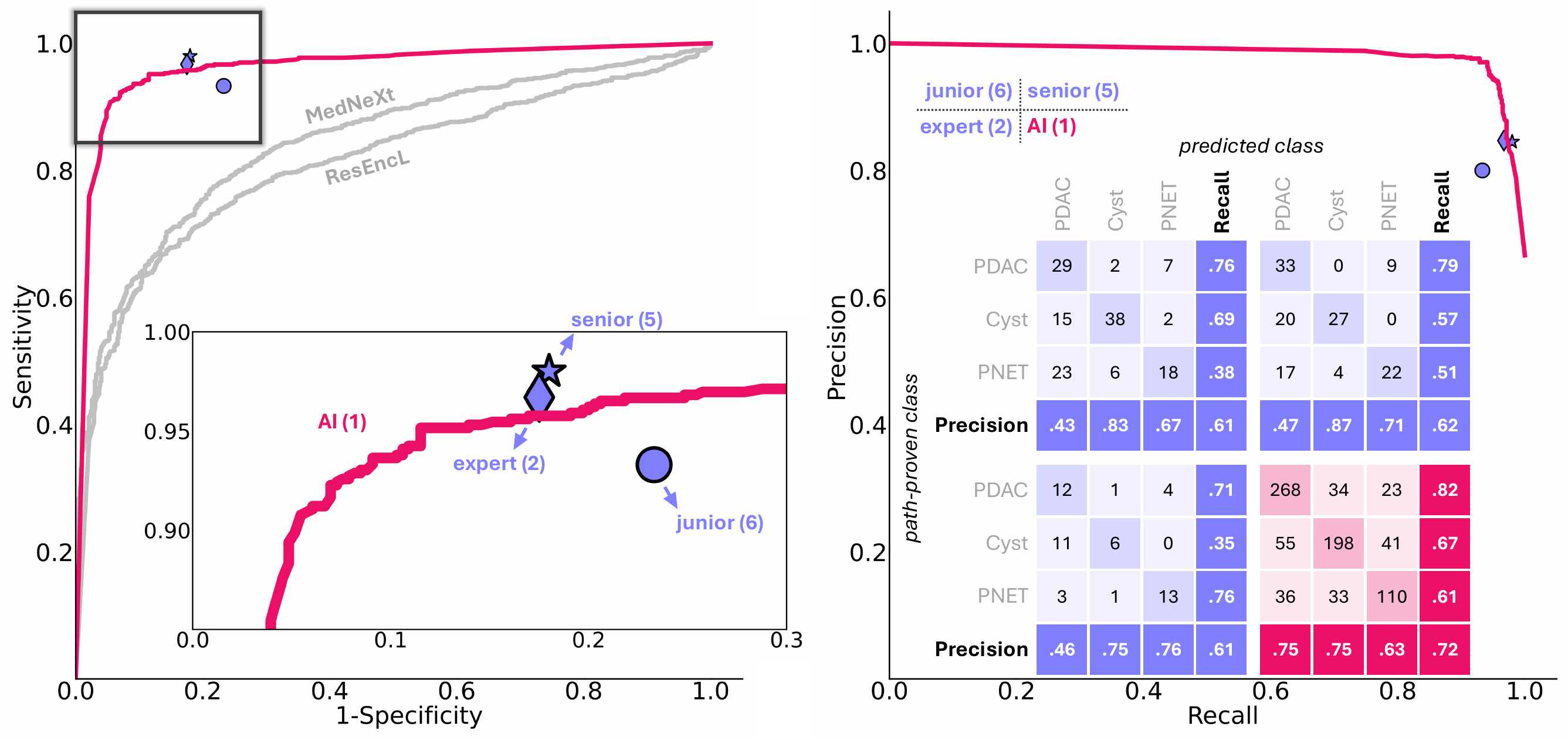}
    \caption{\textbf{\flagship\ matches senior and expert radiologists in tumor detection and surpasses them in tumor diagnosis accuracy.} We conducted an extensive multi-institution, multi-reader study comparing \flagship\ with radiologists with varying levels of experience. Thirteen board-certified radiologists were participated, including 6 juniors ($<$8 years of experience), 5 seniors (8--15 years), and 2 experts ($>$15 years). Each radiologist independently reviewed contrast-enhanced abdominal CT scans in the venous and arterial phases from 50 patients (100 CT scans), representing a broad spectrum of pancreatic conditions, including normal cases and tumors of three common subtypes: cysts, pancreatic adenocarcinoma (PDAC), and pancreatic neuroendocrine tumors (PNET). Radiologists were blinded to the proportion of normal and tumor cases and tasked with detecting and localizing tumors using 3D Slicer by marking any point within the tumor. They also classified tumors into the specified subtypes without access to patient medical history or symptom information. \flagship\ was evaluated under identical conditions on a larger cohort of 982 patients (1,964 CT scans). Performance was assessed using Receiver Operating Characteristic (ROC) and Precision-Recall (PR) curves for tumor detection and confusion matrices for diagnosis. For tumor detection, \flagship\ (\textcolor{flagship}{pink curve}) achieved performance comparable to expert (\textcolor{radiologist}{blue diamond}) and senior radiologists (\textcolor{radiologist}{blue star}), outperforming junior radiologists (\textcolor{radiologist}{blue circle}). In tumor diagnosis of PDAC, cysts, and PNET, \flagship\ achieved 72\% accuracy, exceeding junior, senior, and expert radiologists by 11\%, 10\%, and 11\%, respectively. 
    }
    \label{fig:supp_reader_study}
\end{figure}

\begin{figure}[t]
    \centering
    \includegraphics[width=\linewidth]{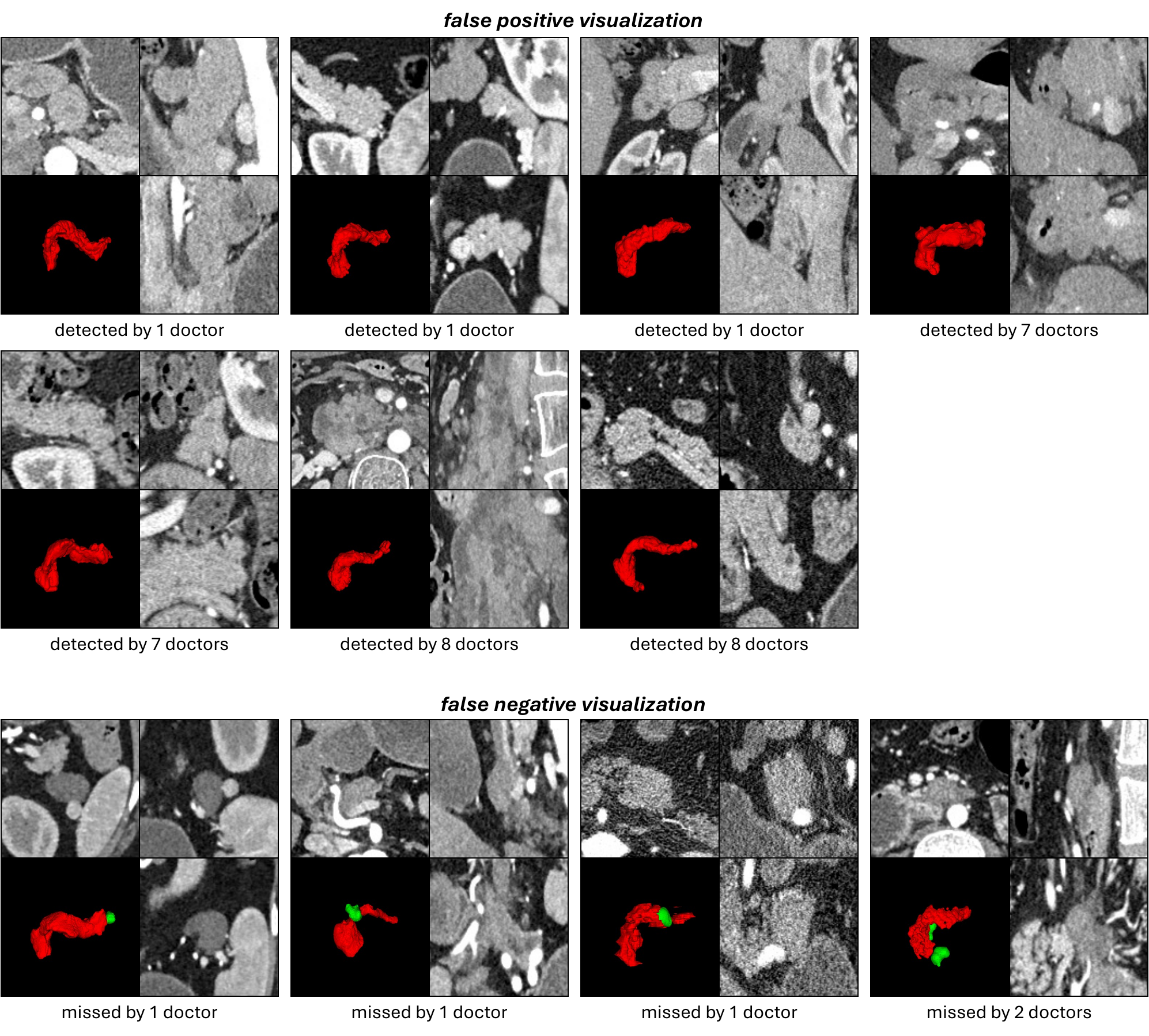}
    \caption{\textbf{Visualization of false positives and negatives predicted by radiologists.} In the false positive cases, the radiologists noticed slight irregularities in the pancreas tissue texture. However, these cases lacked two key reliable warning signs that typically indicate pancreatic tumor: abnormal widening of the main pancreatic duct and localized tissue shrinkage. The false negative cases demonstrated more subtle findings. One case showed a tumor growing outward from the tail end of the pancreas---a location that is often difficult for human readers if not examined thoroughly. In two other cases, while no obvious tumors were visible, there were areas where the pancreas tissue had become unusually thin, which often signals an underlying tumor in that location.}
    \label{fig:supp_visualize_reader_study}
\end{figure}

\begin{table}[t]
\centering
\scriptsize
\caption{\textbf{\flagship\ matches the pancreatic tumor detection performance of senior and expert radiologists, outperforming junior radiologists.} We present the sensitivity (\%) and specificity (\%) of pancreatic tumor detection across different tumor sizes---small ($<$20 mm), medium (20–40 mm), large ($>$40 mm), and all sizes—for radiologists at various career stages and \flagship. 13 radiologists of varying experience levels each evaluated 50 patients individually, while \flagship\ was tested on 982 patients. For all tumor sizes, all radiologist groups demonstrated high sensitivity, but only senior and expert radiologists achieved good specificity. \flagship\ surpassed all radiologist groups in specificity and had higher sensitivity than junior radiologists, approaching the performance of senior and expert radiologists. Notably, \flagship\ attained 100\% sensitivity for medium and large tumors, suggesting it could assist all radiologists in detecting medium-sized tumors and help junior radiologists with large tumor detection. These findings indicate that \flagship\ performs at a level comparable to experienced radiologists, highlighting its potential as a reliable tool in clinical practice.}
\begin{tabular}{p{0.15\linewidth}P{0.1\linewidth} P{0.11\linewidth}P{0.12\linewidth}P{0.12\linewidth}P{0.12\linewidth}P{0.11\linewidth}}
    \toprule
     & & \multicolumn{4}{c}{Sensitivity, \%} & Specificity, \% \\
    \cmidrule(lr){3-6}\cmidrule(lr){7-7}
    career stage & reader & all-size & small ($<$20mm) & medium (20--40mm) & large ($>$40mm) & normal \\
    \midrule
     \multirow{7}{*}{\makecell[l]{junior ($<$8 years)}} 
     & Reader~1 & 96.7{\tiny~(29/30)} & 100{\tiny~(10/10)} & 90.0{\tiny~(9/10)} & 100{\tiny~(10/10)} & 75.0{\tiny~(15/20)} \\
    & Reader~2 & 100{\tiny~(30/30)} & 100{\tiny~(10/10)} & 100{\tiny~(10/10)} & 100{\tiny~(10/10)} & 75.0{\tiny~(15/20)} \\
    & Reader~3 & 100{\tiny~(30/30)} & 100{\tiny~(10/10)} & 100{\tiny~(10/10)} & 100{\tiny~(10/10)} & 80.0{\tiny~(16/20)} \\
    & Reader~4 & 96.7{\tiny~(29/30)} & 100{\tiny~(10/10)} & 90.0{\tiny~(9/10)} & 100{\tiny~(10/10)} & 80.0{\tiny~(16/20)} \\
    & Reader~5 & 76.7{\tiny~(23/30)} & 90.0{\tiny~(9/10)} & 80.0{\tiny~(8/10)} & 60.0{\tiny~(6/10)} & 85.0{\tiny~(17/20)} \\
    & Reader~6 & 90.0{\tiny~(27/30)} & 100{\tiny~(10/10)} & 90.0{\tiny~(9/10)} & 80.0{\tiny~(8/10)} & 65.0{\tiny~(13/20)} \\
    \cmidrule(lr){3-7}
    & average & 93.3{\tiny~(168/180)} & 98.3{\tiny~(59/60)} & 91.7{\tiny~(55/60)} & 90.0{\tiny~(54/60)} & 76.7{\tiny~(92/120)} \\
    \midrule
    \multirow{6}{*}{\makecell[l]{senior (8--15 years)}} 
    & Reader~7 & 96.7{\tiny~(29/30)} & 100{\tiny~(10/10)} & 90.0{\tiny~(9/10)} & 100{\tiny~(10/10)} & 80.0{\tiny~(16/20)} \\
    & Reader~8 & 100{\tiny~(30/30)} & 100{\tiny~(10/10)} & 100{\tiny~(10/10)} & 100{\tiny~(10/10)} & 80.0{\tiny~(16/20)} \\
    & Reader~9 & 96.7{\tiny~(29/30)} & 100{\tiny~(10/10)} & 90.0{\tiny~(9/10)} & 100{\tiny~(10/10)} & 80.0{\tiny~(16/20)} \\
    & Reader~10 & 100{\tiny~(30/30)} & 100{\tiny~(10/10)} & 100{\tiny~(10/10)} & 100{\tiny~(10/10)} & 80.0{\tiny~(16/20)} \\
    & Reader~11 & 96.7{\tiny~(29/30)} & 100{\tiny~(10/10)} & 90.0{\tiny~(9/10)} & 100{\tiny~(10/10)} & 90.0{\tiny~(18/20)} \\
    \cmidrule(lr){3-7}
    & average & 98.0{\tiny~(147/150)} & 100{\tiny~(50/50)} & 94.0{\tiny~(47/50)} & 100{\tiny~(50/50)} & 82.0{\tiny~(82/100)} \\
    \midrule
    \multirow{3}{*}{\makecell[l]{expert ($>$15 years)}}
    & Reader~12  & 93.3{\tiny~(28/30)} & 100{\tiny~(10/10)} & 80.0{\tiny~(8/10)} & 100{\tiny~(10/10)} & 80.0{\tiny~(16/20)} \\
    & Reader~13  & 100{\tiny~(30/30)} & 100{\tiny~(10/10)} & 100{\tiny~(10/10)} & 100{\tiny~(10/10)} & 85.0{\tiny~(17/20)} \\
    \cmidrule(lr){3-7}
    & average & 96.7{\tiny~(58/60)} & 100{\tiny~(20/20)} & 90.0{\tiny~(18/20)} & 100{\tiny~(20/20)} & 82.5{\tiny~(33/40)} \\
    \midrule
     & \flagship & 94.1{\tiny~(640/680)} & 85.2{\tiny~(468/549)} & 100{\tiny~(105/105)} & 100{\tiny~(10/10)} & 83.8{\tiny~(253/302)} \\
    \bottomrule
\end{tabular}
\label{tab:supp_reader_study}
\end{table}

\clearpage
\subsection{\textbf{\dataset} vs. Public Organ and Tumor Datasets}\label{sec:supp_related_datasets}

\begin{table*}[h]
\centering
\scriptsize
\caption{
\textbf{Comparison of public organ and pancreatic tumor CT datasets.}
\dataset\ surpasses prior organ and pancreatic CT datasets in both scale and completeness, spanning \numofct\ CTs across \numofhospital\ centers and \numofcountry\ countries. It integrates heterogeneous scanners, full demographic metadata (age, sex, phase), and clinically meaningful labels (narrative and structured reports), forming the most comprehensive foundation for scalable medical intelligence to date.
}
\begin{tabular}{
p{0.20\linewidth}  
P{0.03\linewidth}  
P{0.05\linewidth}  
P{0.05\linewidth}  
P{0.07\linewidth}  
P{0.05\linewidth}  
P{0.05\linewidth}  
P{0.02\linewidth}  
P{0.03\linewidth}  
P{0.04\linewidth}  
P{0.06\linewidth}  
P{0.06\linewidth}  
}
\toprule
& \multicolumn{6}{c}{\textbf{Scale}} & \multicolumn{3}{c}{\textbf{Metadata}} & \multicolumn{2}{c}{\textbf{Clinical}} \\
\cmidrule(lr){2-7} \cmidrule(lr){8-10} \cmidrule(lr){11-12}
\textbf{Dataset} & 
\makecell{\# of\\CT} &
\makecell{\# of\\new CT} &
\makecell{\# of\\structure} &
\makecell{\# of\\annotation} &
\makecell{\# of\\center} &
\makecell{\# of\\country} &
\makecell{age} &
\makecell{sex} &
\makecell{phase} &
\makecell{narrative\\report} &
\makecell{structured\\report} \\
\midrule
CHAOS~\citeyearpar{valindria2018multi} & 40 & 40 & 1 & 40 & 1 & 1 & \no & \no & \no & \no & \no \\
BTCV~\citeyearpar{landman2015miccai} & 50 & 50 & 12 & 600 & 1 & 1 & \no & \no & \no & \no & \no \\
CT-ORG~\citeyearpar{rister2020ct} & 140 & 140 & 6 & 840 & 8 & 6 & \no & \no & \no & \no & \no \\
WORD~\citeyearpar{luo2021word} & 150 & 150 & 16 & 2400 & 1 & 1 & \no & \no & \no & \no & \no \\
LiTS~\citeyearpar{bilic2019liver} & 201 & 201 & 1 & 201 & 7 & 5 & \no & \no & \no & \no & \no \\
AMOS22~\citeyearpar{ji2022amos} & 500 & 500 & 15 & 7500 & 2 & 1 & \no & \no & \no & \no & \no \\
KiTS~\citeyearpar{heller2019kits19} & 599 & 599 & 1 & 599 & 1 & 1 & \no & \no & \no & \no & \no \\
AbdomenCT-1K~\citeyearpar{ma2021abdomenct} & 1,112 & \no & 4 & 4448 & 12 & 7 & \no & \no & \no & \no & \no \\ 
TotalSegmentator~\citeyearpar{wasserthal2023totalsegmentator} & 1,228 & 1,228 & 117 & 143,676 & 10 & 1 & \yes & \yes & \yes & \no & \no \\
FLARE'23~\citeyearpar{ma2024automatic} & 4,100 & \no & 13 & 53,300 & 30 & \no & \no & \no & \no & \no & \no \\
Trauma Detect.~\citeyearpar{rudie2024rsna} & 4,711 & 4,711 & \no & \no & 23 & 12 & \no & \no & \no & \no & \no \\
AbdomenAtlas~\citeyearpar{li2024well,chen2025scaling,bassi2025radgpt} & 9,262 & \no & 25 & 231,550 & 89 & 17 & \yes & \yes & \yes & 9,262 & 9,262 \\
\midrule
TCIA–panNET~\citeyearpar{chen2023special} & 38 & 38 & 1 & 38 & 1 & 1 & \yes & \yes & \yes & \no & \no \\
TCIA-Pancreas~\citeyearpar{han2021deep} & 82 & 82 & 1 & 82 & 1 & 1 & \no & \no & \yes & \no & \no \\
MSD–Pancreas~\citeyearpar{antonelli2022medical} & 420 & 420 & 1 & 420 & 1 & 1 & \no & \no & \yes & \no & \no \\
PANORAMA~\citeyearpar{alves2024panorama} & 2,238 & 1,964 & 5 & 11,190 & 5 & 3 & \yes & \yes & \no & \no & \no \\
PanTS~\citeyearpar{li2025pants} & 9,901 & 556 & 27 & 267,327 & 76 & 12 & \yes & \yes & \yes & \no & 9,901 \\
\midrule
\dataset\ (Ours) & \textbf{\numofct} & \textbf{\numofnewct} & \textbf{\numofclass} & \textbf{\numofmask} & \textbf{\numofhospital} & \textbf{\numofcountry} &
\yes & \yes & \yes & \textbf{\numofct} & \textbf{\numofct} \\
\bottomrule
\end{tabular}
\label{tab:dataset_comparison}
\end{table*}

\clearpage
\section{Experimental Results of \flagship}\label{sec:supp_result}

\subsection{Benchmarking \textbf{\flagship}: Evaluation Metrics}\label{sec:supp_evaluation_metrics}
 % \TODO{Add function for Sensitivity, specificty, and F1-score in supplementary. Add DSC and NSD formula in supplementary.} 
 % Diagnosis Metrics

\smallskip\noindent\textbf{\textit{Diagnosis}}. For the tumor detection task in the binary classification setting, we use Sensitivity, Specificity and F1-score as our evaluation metrics. Let \text{TP}, \text{TN}, \text{FP}, and \text{FN} denote the numbers of true positives, true negatives, false positives, and false negatives, respectively. We define sensitivity (also called recall) as the fraction of actual positives that are correctly identified:
\begin{equation}
\text{Sensitivity} = \frac{\text{TP}}{\text{TP} + \text{FN}}.
\end{equation}
Specificity measures the fraction of actual negatives that are correctly identified:
\begin{equation}
\text{Specificity} = \frac{\text{TN}}{\text{TN} + \text{FP}}.
\end{equation}
The F1-score\footnote{The F1-score combines sensitivity (recall) and precision, providing a balanced metric for scenarios involving class imbalance.} combines precision ($\text{Precision} = \frac{\text{TP}}{\text{TP} + \text{FP}}$) and recall ($\text{Recall} = \frac{\text{TP}}{\text{TP} + \text{FN}}$) into a single metric, and is defined as:
\begin{equation}
\text{F1-score} = 2 \times \frac{\text{Precision} \times \text{Recall}}{\text{Precision} + \text{Recall}}.
\end{equation}

% Segmentation Metrics

\smallskip\noindent\textbf{\textit{Segmentation}}. For the tumor segmentation task, we use the Dice Similarity Coefficient (DSC) and Normalized Surface Dice (NSD) as our evaluation metrics. The DSC compares the overlap between two sets, commonly the predicted segmentation A and the ground-truth segmentation B. It is given by:
\begin{equation}
\text{DSC} = \frac{2 \lvert A \cap B \rvert}{\lvert A \rvert + \lvert B \rvert}.
\end{equation}
The NSD metric quantifies how closely the surfaces of the predicted and ground-truth segments match within a specified tolerance $\delta$. Let $S_{p}$ and $S_{g}$ denote the sets of boundary points for the predicted and ground-truth surfaces, respectively, and let $d(x, S)$ represent the minimum distance from a point x to any point in set S. The NSD is then defined as the proportion of boundary points (from both $S_{p}$ and $S_{g}$) that lie within $\delta$ of each other:
\begin{equation}
\text{NSD} = \frac{\sum_{x \in S_{p}} \mathbf{1}[\,d(x, S_{g}) < \delta] \;+\; \sum_{x \in S_{g}} \mathbf{1}[\,d(x, S_{p}) < \delta]}{\lvert S_{p} \rvert + \lvert S_{g} \rvert},
\end{equation}
where $\mathbf{1}[\cdot]$ is the indicator function, equal to 1 if the condition is satisfied and 0 otherwise.

\clearpage
\subsection{Benchmarking \textbf{\flagship}: Dataset Attributes}\label{sec:supp_dataset_attributes}

\begin{figure}[h]
\centering  
    \includegraphics[width=1\linewidth]{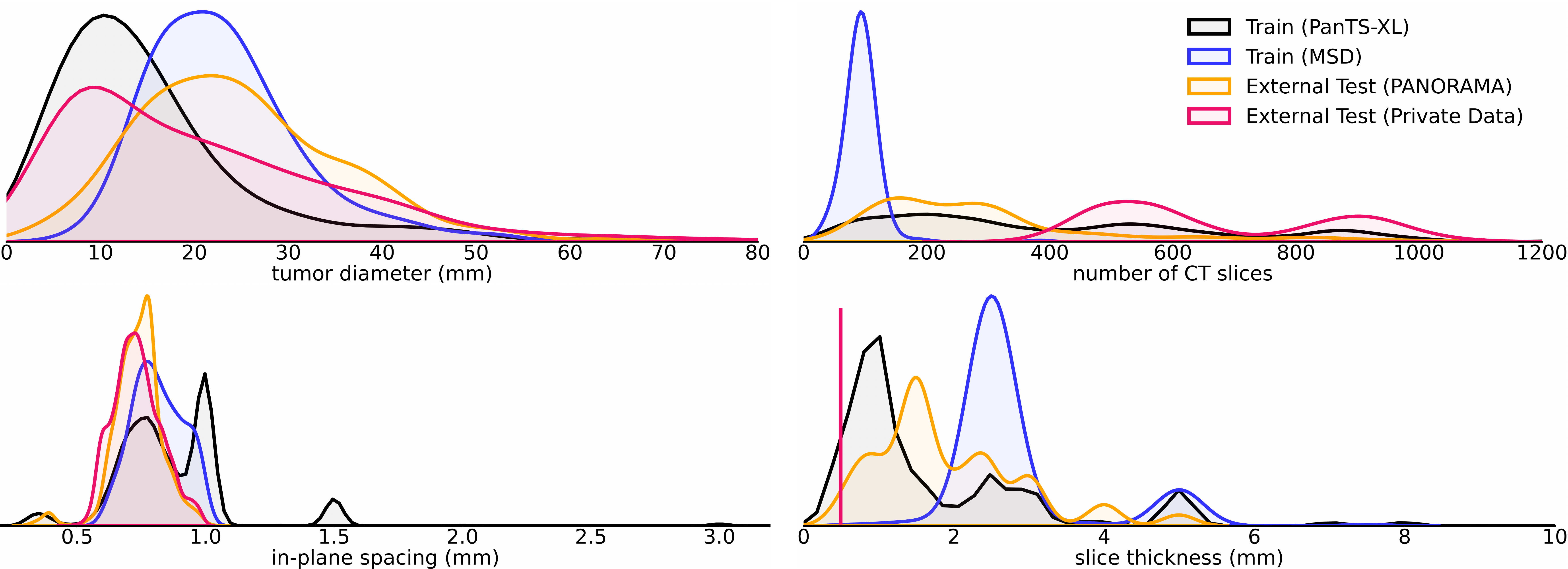}
    \caption{\textbf{Dataset attributes.}
    Our study used four datasets: two for training AI models and two for testing them. Before the creation of \dataset\ (\numofct\ CT scans), the publicly available MSD-Pancreas dataset was the \textit{only} resource for training pancreatic tumor segmentation models. Therefore, all baseline AI models in this study were trained on MSD-Pancreas. Existing literature suggests that AI is vulnerable when applied to CT scans from datasets with differing attributes \cite{shin2016deep, guan2021domain}, such as variations in tumor diameter, the number of CT slices, in-plane spacing, and slice thickness. To evaluate the robustness of the baseline models, we conducted external validation using two datasets sourced from hospitals distinct from those contributing to MSD-Pancreas (Memorial Sloan Kettering Cancer Center, USA). The first external dataset, PANORAMA, was sourced from five hospitals across three countries, including Dutch, Sweden and Norway \cite{alves2024panorama}. The second dataset, a proprietary dataset, was gathered from hospitals in a country distinct from MSD-Pancreas. As seen, compared with MSD-Pancreas, the PANORAMA dataset differs in the number of CT slices and slice thickness, while the proprietary dataset diverges significantly across all four attributes. The test results in \tableautorefname~\ref{tab:combined_detection_segmentation} reveal that baseline models perform better on the PANORAMA, with the proprietary dataset yielding a much lower performance. These findings validate the hypothesis that discrepancies between training and test data significantly impact AI performance and robustness. This motivated us to create \dataset\ that offers a substantially larger training set (47,000 annotated CT scans~vs.~MSD-Pancreas's 200) with greater diversity in key attributes as illustrated by the black curve.
    }
    \label{fig:supp_metadata_histogram}
\end{figure}

\clearpage

\subsection{PDAC, Cyst, and PNET Diagnosis (+7\% Accuracy)}
\label{sec:supp_tumor_classification}

\begin{figure}[h]
    \centering
    \includegraphics[width=1\linewidth]{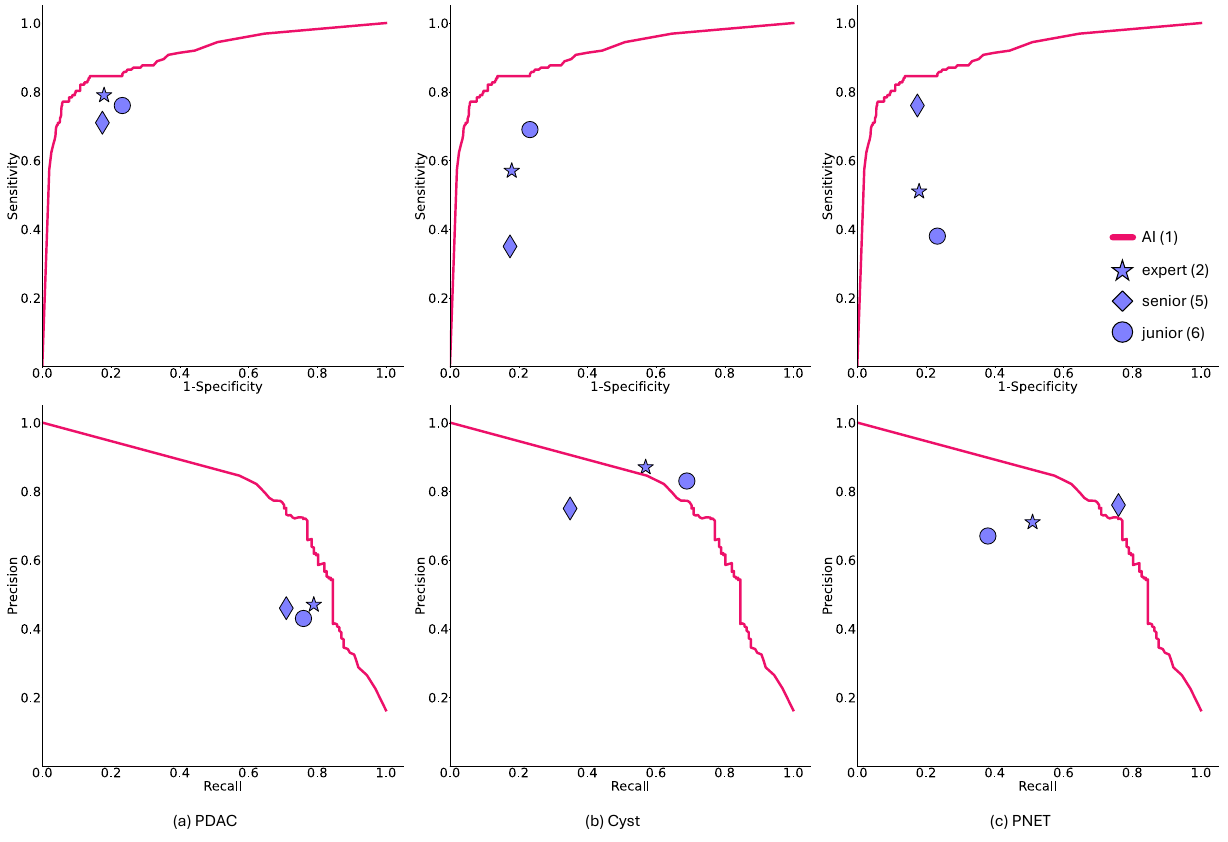}
    \caption{
    \textbf{Our \flagship\ can approach radiologists' performance in pancreatic tumor diagnosis.} 
Classifying pancreatic tumor types (Cyst, PDAC, PNET) directly from CT scans is challenging due to the subtle and overlapping visual features among these tumors. Key characteristics such as shape, size, and enhancement patterns can vary significantly within the same tumor type and often mimic those of other types. Additionally, CT scans lack the biological and molecular context provided by patient symptoms, medical history, follow-up imaging, or biopsy results, which are crucial for accurate diagnosis. Furthermore, variations in imaging protocols and scanner settings across institutions add complexity, making it difficult for both radiologists and AI models to achieve high accuracy. Using our annotated dataset of tumor types, this study marks the first time that: (1) AI performance is evaluated on a publicly available dataset, enabling reproducibility. (2) Radiologists are tested on the same dataset, allowing others to benchmark their performance. (3) AI is directly compared with radiologists across different career stages.
    }    
    \label{fig:supp_tumor_classification}
\end{figure}

\clearpage

\subsection{Pancreatic Tumor Detection (+10\% Sensitivity)}
\label{sec:supp_tumor_detection}

\begin{table*}[h]
\centering
\scriptsize
\caption{\textbf{\flagship, with a backbone of ResEncL, achieves the best performance for pancreatic tumor detection.} Note that these are tumor-wise detection results. Performance is given as sensitivity, specificity, and F1-score. Best-performing results are \textbf{bolded} for each dataset. In addition, we have performed a one-sided Wilcoxon signed rank test between the best-performing model and others~\cite{wiesenfarth2021methods}. The performance gain is statistically significant at the $P=0.05$ level, with highlighting in a \textcolor{flagship!75}{pink} box.
}
\begin{tabular}{p{0.18\linewidth}p{0.12\linewidth}P{0.1\linewidth}
P{0.1\linewidth}
P{0.1\linewidth}
P{0.1\linewidth}}
    \toprule
     &  & \multicolumn{1}{c}{PANORAMA ($N$=1,964)$^\dag$} & \multicolumn{3}{c}{Proprietary dataset ($N$=1,958)}\\
    \cmidrule(lr){3-3}\cmidrule(lr){4-6}
    method & training set 
    & Sensitivity 
    & Sensitivity & Specificity & F1-score \\
    \midrule

    Swin UNETR~\cite{tang2022self} & MSD-Pancreas 
    & 85.5{\tiny~(497/581)}
    & 25.2{\tiny~(824/3426)} & 16.7{\tiny~(104/623)} & 35.9{\tiny~(1728/4809)} \\

    UniSeg~\cite{ye2023uniseg} & MSD-Pancreas 
    & 77.8{\tiny~(452/581)}
    & 23.4{\tiny~(801/3426)} & 78.5{\tiny~(485/623)} & 36.7{\tiny~(1602/4361)} \\ 

    ResEncL~\cite{isensee2024nnu} & MSD-Pancreas
    & 74.7{\tiny~(434/581)}
    & 23.7{\tiny~(813/3426)} & 87.0{\tiny~(542/623)} & 38.1{\tiny~(1660/4360)} \\ 

    STU-Net-Base~\cite{huang2023stu} & MSD-Pancreas
    & 76.8{\tiny~(446/581)}
    & 23.1{\tiny~(791/3426)} & 84.4{\tiny~(526/623)} & 36.7{\tiny~(1582/4314)} \\ 

    MedNeXt~\cite{roy2023mednext} & MSD-Pancreas
    & 73.3{\tiny~(426/581)}
    & 24.6{\tiny~(842/3426)} & 85.2{\tiny~(531/623)} & 39.3{\tiny~(1726/4394)} \\ 

    DTI~\cite{liu2025ai} & PANORAMA
    & --
    & 26.7\tiny~(916/3426)
    & \cellcolor{flagship!12}88.1\tiny~(549/623)
    & 41.4\tiny~(1832/4416)\\

    \midrule
    \flagship & \dataset
    & \cellcolor{flagship!12}\textbf{89.2{\tiny~(518/581)}}
    & \cellcolor{flagship!12}\textbf{40.2{\tiny~(1381/3426)}} & \cellcolor{flagship!12}\textbf{88.3{\tiny~(550/623)}} & \cellcolor{flagship!12}\textbf{56.6{\tiny~(2754/4865)}}\\
    
    $\Delta$ & 
    & \textbf{\textcolor{red}{+3.7}}
    & \textbf{\textcolor{red}{+13.5}}
    & \textbf{\textcolor{red}{+0.2}}
    & \textbf{\textcolor{red}{+17.3}}
    \\

    \bottomrule
\end{tabular}
\begin{tablenotes}
    \item $^\dag$PANORAMA annotates only PDAC, treating all other types of pancreatic tumors and healthy pancreases as \textit{Normal}---specificity and F1-score cannot be computed.
\end{tablenotes}
\label{tab:supp_tumor_wise_detection}
\end{table*}

\begin{figure}[h]
    \centering
    \includegraphics[width=1\linewidth]{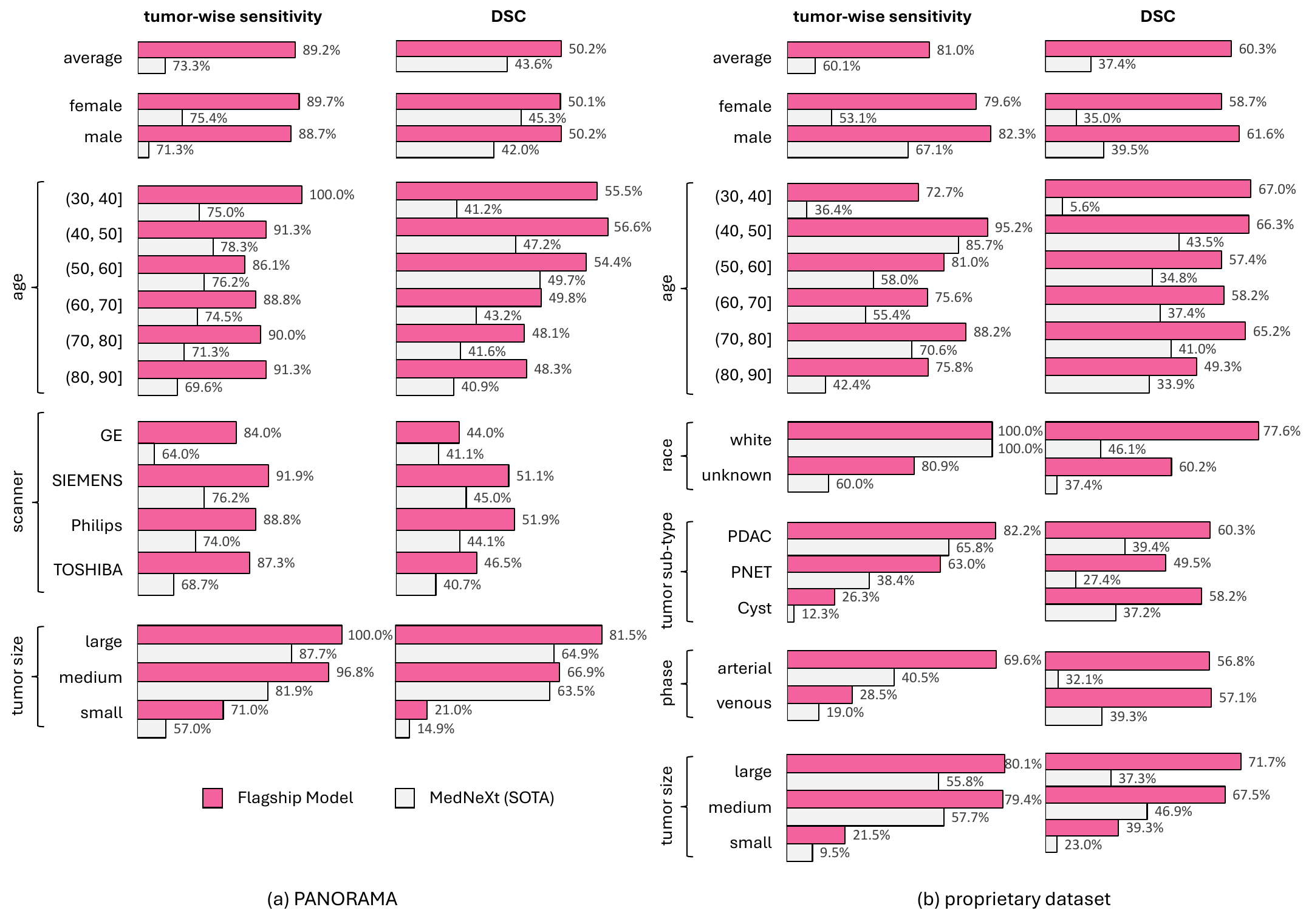}
    \caption{\textbf{\flagship\ demonstrates robust generalizability across diverse demographic and technical variations in out-of-distribution evaluations.} \flagship\ shows improved tumor detection and segmentation performance across various age, sex, race, tumor sub-types, tumor sizes, scanner types, and imaging phases. \flagship, trained on \dataset\ (large-scale but silver standard), consistently matches or surpasses the performance of the MedNeXt model (top 1), which was trained on a smaller but gold standard dataset.
    }    
    \label{fig:supp_tumor_detection_metadata_analysis}
\end{figure}

\clearpage
\subsection{Pancreatic Tumor Segmentation (+14\% DSC)}
\label{sec:supp_tumor_segmentation}

\begin{table*}[h]
\centering
\scriptsize
\caption{\textbf{\flagship\ demonstrates robust generalizability across various demographic groups and scanner types in tumor segmentation.} We compare the median DSC and interquartile range (IQR) of \flagship\ and public top-performing MedNeXt model on the PANORAMA and proprietary datasets for sex, age, scanner type, and race. Notably, \flagship\ consistently achieves higher median DSC with statistically significant improvements (p-value$<$0.001 in most cases). For example, in the age group 70–80, \flagship\ achieved a median DSC of 73.9\% compared to MedNeXt's 47.8\%, a difference of 16.1\% (p-value$<$0.001).}
\begin{tabular}{p{0.001\linewidth}p{0.065\linewidth}P{0.11\linewidth}P{0.11\linewidth}P{0.08\linewidth}P{0.045\linewidth}P{0.11\linewidth}P{0.11\linewidth}P{0.08\linewidth}P{0.045\linewidth}}
    \toprule
     & & \multicolumn{2}{c}{Median DSC (IQR) on PANORAMA, \%} & & & \multicolumn{2}{c}{Median DSC (IQR) on proprietary dataset, \%}  \\
    \cmidrule(lr){3-4}\cmidrule(lr){7-8}
    \multicolumn{2}{l}{group} & \flagship & MedNeXt (SOTA) & difference & p-value & \flagship & MedNeXt (SOTA) & difference & p-value \\
    \midrule
    \multicolumn{2}{l}{all test samples} & 58.1\tiny{ (24.4--76.2)} & 54.4{\tiny~(0.0--74.8)} & 0.1{\tiny~(-11.2--20.1)} & $<$0.001 & 70.4{\tiny~(44.4--81.3)} & 38.8{\tiny~(0.0--67.7)} & 15.2{\tiny~(2.7--40.8)} & $<$0.001\\
    \multicolumn{2}{l}{sex} \\
     & female & 57.5{\tiny~(25.0--77.0)} & 55.9{\tiny~(0.6--76.3)} & 0.0{\tiny~(-12.7--18.7)} & {0.072} & 68.8{\tiny~(39.8--79.8)} & 31.4{\tiny~(0.0--68.0)} & 14.0{\tiny~(1.5--41.2)} & $<$0.001 \\
     & male & 58.9{\tiny~(23.9--74.8)} & 53.4{\tiny~(0.0--72.9)} & 0.7{\tiny~(-9.8--21.0)} & 0.002 & 71.4{\tiny~(49.2--82.5)} & 45.7{\tiny~(0.3--67.5)} & 15.5{\tiny~(3.5--40.1)} & $<$0.001 \\
    \multicolumn{2}{l}{age} \\
     & 30--40 & 66.6{\tiny~(40.4--81.7)} & 44.0{\tiny~(12.9--72.3)} & 4.9{\tiny~(-1.4--20.6)} & 0.608 & 70.5{\tiny~(63.6--73.2)} & 0.1{\tiny~(0.0--12.5)} & 62.8{\tiny~(54.6--69.9)} & $<$0.001 \\
     & 40--50 & 66.9{\tiny~(52.9--82.3)} & 57.4{\tiny~(8.8--74.2)} & 0.6{\tiny~(-2.8--15.8)} & 0.325 & 74.8{\tiny~(65.0--84.0)} & 54.0{\tiny~(16.3--68.9)} & 17.7{\tiny~(7.6--58.2)} & 0.012 \\
     & 50--60 & 63.5{\tiny~(31.1--80.2)} & 60.0{\tiny~(16.9--77.9)} & 0.0{\tiny~(-14.6--19.2)} & 0.299 & 65.4{\tiny~(42.1--78.6)} & 36.5{\tiny~(0.0--62.7)} & 16.0{\tiny~(1.1--42.4)} & $<$0.001 \\
     & 60--70 & 57.5{\tiny~(26.2--75.9)} & 54.1{\tiny~(0.0--71.9)} & 0.0{\tiny~(-11.7--20.8)} & 0.038 & 68.0{\tiny~(38.0--81.5)} & 41.1{\tiny~(0.0--66.8)} & 13.8{\tiny~(3.1--35.7)} & $<$0.001  \\
     & 70--80 & 55.8{\tiny~(21.6--74.5)} & 50.1{\tiny~(0.0--74.4)} & 0.0{\tiny~(-11.1--19.2)} & 0.039 & 73.9{\tiny~(57.0--83.1)} & 47.8{\tiny~(0.7--71.9)} & 16.1{\tiny~(3.8--40.6)} & $<$0.001  \\
     & 80--90 & 55.4{\tiny~(24.2--70.7)} & 44.2{\tiny~(0.0--75.4)} & 0.4{\tiny~(-6.3--25.0)} & 0.263 & 57.3{\tiny~(9.3--78.7)} & 31.7{\tiny~(0.0--66.8)} & 8.2{\tiny~(0.0--18.7)} & $<$0.001  \\
    \multicolumn{2}{l}{scanner} \\
     & GE & 47.1{\tiny~(20.4--69.4)} & 54.5{\tiny~(0.0--74.2)} & 0.0{\tiny~(-4.1--10.0)} & 0.754 & - & - & - & -\\
     % & Canon & 70.2{\tiny~(54.5--84.1)} & 54.0{\tiny~(27.3--76.0)} & 13.4{\tiny~(1.9--16.5)} & 0.582 & - & - & - & -\\
     & SIEMENS & 57.7{\tiny~(26.4--76.9)} & 57.3{\tiny~(0.8--75.7)} & 0.5{\tiny~(-12.9--21.9)} & 0.072 & - & - & - & -\\
     & Philips & 59.8{\tiny~(30.0--75.9)} & 55.4{\tiny~(0.0--72.0)} & 0.2{\tiny~(-11.6--22.4)} & 0.006 & - & - & - & -\\
     & TOSHIBA & 57.3{\tiny~(16.1--75.4)} & 44.1{\tiny~(0.0--75.5)} & 0.0{\tiny~(-8.3--17.6)} & 0.157 & - & - & - & -\\
    \multicolumn{2}{l}{race} \\
     & white  & - & - & - & - & 77.6{\tiny~(76.6--78.5)} & 46.1{\tiny~(43.0--49.3)} & 31.4{\tiny~(29.3--33.6)} & 0.102\\
     & unknown  & - & - & - & - & 70.3{\tiny~(44.0--81.3)} & 38.6{\tiny~(0.0--67.8)} & 15.0{\tiny~(2.5--40.9)} & $<$0.001 \\
    \bottomrule
\end{tabular}
\label{tab:metadata_analysis_tumor_segmentation}
\end{table*}

\end{document}